\documentclass[journal]{IEEEtran}

\usepackage{color}
\usepackage{algorithm}
\usepackage{algorithmic}
\usepackage{amsmath} 
\usepackage{amsthm}
\usepackage{amssymb}
\usepackage{graphicx}
\usepackage{epstopdf}
\usepackage{booktabs}
\usepackage{threeparttable}
\usepackage{multirow}
\usepackage{tabularx}
\usepackage{cite}
\usepackage{mathrsfs}

\newtheorem{mydef}{Definition}
\newtheorem{mytheo}{Theorem}

\usepackage[normalem]{ulem}
\useunder{\uline}{\ul}{}

\title{Coarse to Fine Two-Stage Approach to Robust Tensor Completion of Visual Data
\thanks{This work was supported by NSF CAREER Award CCF-1552497 and NSF Award CCF-2106339.

$*$ Department of Electrical and Computer
Engineering, University of Central Florida, Orlando, FL, 32816, USA.

$\dagger$ Department of Computer
Science, University of Central Florida, Orlando, FL, 32816, USA.

E-mails: \{Yicong.He, George.Atia\}@ucf.edu. 
}}

\author{Yicong He$^*$ and George K. Atia$^{*,\dagger}$, \IEEEmembership{Senior~Member,~IEEE}}

\begin{document}

\maketitle

\begin{abstract}
Tensor completion is the problem of estimating the missing values of high-order data from partially observed entries. Data corruption due to prevailing outliers poses major challenges to traditional tensor completion algorithms, which catalyzed the development of robust algorithms that alleviate the effect of outliers. However, existing robust methods largely presume that the corruption is sparse, which may not hold in practice. In this paper, we develop a two-stage robust tensor completion approach to deal with tensor completion of visual data with a large amount of gross corruption. A novel coarse-to-fine framework is proposed which uses a global coarse completion result to guide a local patch refinement process. To efficiently mitigate the effect of a large number of outliers on tensor recovery, we develop a new M-estimator-based robust tensor ring recovery method which can {adaptively} identify the outliers and alleviate their negative effect in the optimization. The experimental results demonstrate the superior performance of the proposed approach over state-of-the-art robust algorithms for tensor completion.
\end{abstract}


\section{Introduction}
\label{Intro}
Predicting missing information from partially observed data is an emerging topic in modern data science due to unprecedented growth in data volume and dimensionality \cite{song2019tensor}. In multi-way data analysis where the data can be represented as a high-order tensor, this problem can be formulated in the lens of tensor completion with the goal of recovering the missing entries of a partially observed tensor. While the tensor completion problem is ill-posed without further model assumptions, actual formulations exploit the {low-rank} structure intrinsic to much of the real world data. 
{To date, numerous tensor completion algorithms have been proposed based on different tensor decomposition models \cite{liu2014trace,karlsson2016parallel,liu2012tensor,fan2021multi,zhang2016exact,zhou2017tensor,wang2017efficient,huang2020provable,yu2019tensor,bengua2017efficient}, and have been successfully applied to a wide range of problems in computer vision \cite{dian2017hyperspectral,zhang2019computational}, pattern recognition \cite{xie2020robust,tang2021one}, and signal processing\cite{cichocki2015tensor,tang2019tensor}.}

In real applications, data may be corrupted by outliers due to human error or signal interference, making some of the observed data unreliable {\cite{xie2016recover,wang2020anomaly}}. Traditional tensor completion algorithms are largely based on a second-order measure of the error residuals, thus their performance degrades in {the} presence of outliers. {In recent years, many works have focused on robust tensor completion and proposed several algorithms that were shown to outperform traditional completion algorithms in the presence of sparse outliers \cite{goldfarb2014robust,yang2016iterative,huang2020robust,jiang2019robust,liu2021simulated,li2021robust}.}
Despite their robust performance with outlier-corrupted data compared with traditional methods, the usefulness of these algorithms is limited to settings with a small fraction of outliers. When the number of outliers increases, there are primarily two interrelated difficulties. First, the large number of outliers could overwhelm their underlying outlier rejection mechanism, leading to severe performance degradation. For example, when the corruption is non-sparse, the $\ell_1$-norm, which is at the heart of $\ell_1$-norm-based robust tensor completion methods, falls short of accurately capturing the error residual. Second, the percentage of data entries to be relied on for completion of the missing entries decreases accordingly. For example, if $50\%$ of the entries of a given tensor are observed, of which $60\%$ are perturbed with outliers, then only $20\%$ of the entries of the whole tensor are correctly observed. The reduced amount of reliable information for completion renders the tensor completion task more challenging, necessitating different means {of} completion.

To deal with tensor completion in the presence of a large number of outliers, we develop a novel two-stage coarse-to-fine framework for robust tensor completion. At the global coarse stage, a robust tensor completion algorithm is applied to the entire tensor to get a coarse completion result and identify a large number of outliers. At the local refinement stage, for each patch of the given tensor, a novel patch jitter procedure is proposed and used to construct a patch tensor using neighboring patches. Subsequently, robust tensor recovery incorporating the global coarse completion information is performed on the patch tensor, resulting in refined patch tensor recovery. In sharp contrast to existing non-local patch-based tensor completion algorithms \cite{xie2018tensor,zhang2019nonlocal,ding2021tensor}, the proposed patch-based method does not perform block-matching, which greatly saves the computational cost and also avoids biased matching caused by outliers.

Further, to improve both the robustness and completion/refinement performance, we propose a new robust tensor ring recovery algorithm utilizing an M-estimator as the error measure. Tensor ring (TR) rank model has shown desirable performance in many tensor completion tasks owing to its flexibility \cite{zhao2016tensor,wang2017efficient}. M-estimators rooted in robust statistics are generalizations of maximum likelihood (ML) estimators for which the objective function is a sample average \cite{tyler1987distribution}. The selection of a proper loss function for M-estimators can greatly enhance robustness against large outliers. In order to handle the complex objective resulting from the use of an M-estimator, we leverage a half-quadratic (HQ) \cite{nikolova2005analysis} minimization approach whereby the problem is reformulated as a reweighted tensor ring completion program. Then, based on a TR unfolding scheme \cite{yu2019tensor,huang2020provable}, we develop a robust tensor ring completion algorithm utilizing truncated singular value decomposition (SVD) to capture the {low-rank} structure. The proposed robust algorithms are efficient and have a simple structure owing to the use of an HQ-based method and a TR unfolding scheme, and can be applied to both the global tensor completion and local patch refinement stages. Further, the convergence of the proposed algorithm is analyzed. The following summarizes the main contributions of the paper.

\begin{enumerate}
\item We propose a novel two-stage coarse-to-fine framework for robust tensor completion of visual data. First, we perform global coarse completion. Then, local patch refinement is applied to patch tensors created using patch jitter, where prior information from the global coarse completion result is incorporated.

\item We propose a new M-estimator-based tensor ring recovery method for both global tensor completion and local patch refinement. A half-quadratic approach is introduced to transform the non-convex optimization problem to a re-weighted tensor completion problem. Then, a new algorithm is developed based on a TR unfolding scheme and truncated SVD, and its convergence is analyzed.

\item We perform experiments on real data for image and video completion, demonstrating the superior performance of the proposed algorithm compared with existing robust tensor completion algorithms.
\end{enumerate}

The paper is organized as follows. In section \ref{sec:relatedwork}, we present the related work in matrix and tensor completion. In Section \ref{sec:background}, we briefly introduce our notation and provide some preliminary background on tensor ring decomposition and completion. In Section \ref{sec:framework}, we present our two-stage coarse-to-fine tensor completion framework, along with the formulation of the objective function for each stage. In Section \ref{sec:HQTRC}, we propose our new HQ-based robust tensor ring recovery algorithm. Experimental results are presented in Section \ref{sec:results} to demonstrate the completion performance. Finally, the conclusion is given in Section \ref{sec:conc}.
\smallbreak

{
\section{Related work}
\label{sec:relatedwork}

\noindent\textbf{Low-rank matrix and tensor completion.}
Matrix or tensor completion aim to fill the missing entries of a partially observed matrix or tensor data. The key idea underlying the ability to estimate their unknown entries is the low rank property inherent in many machine learning problems \cite{chen2012low,kang2016top,yu2017multi,fan2021robust}, which captures the redundancy and correlation within a matrix or tensor \cite{candes2009exact}.}

{

Different from the matrix domain where the rank is uniquely defined, the rank of a tensor has various definitions corresponding to different tensor factorization (decomposition) models, such as CANDECOMP/PARAFAC (CP) \cite{harshman1970foundations}, Tucker \cite{tucker1966some}, tensor SVD (t-SVD) \cite{kilmer2011factorization}, tensor ring (TR) \cite{zhao2016tensor} and tensor train (TT)  \cite{oseledets2011tensor}. Based on different tensor decomposition models, a large number of tensor completion algorithms were developed \cite{liu2012tensor,liu2014trace,karlsson2016parallel,fan2021multi,zhang2016exact,zhou2017tensor,wang2017efficient,huang2020provable,yu2019tensor,bengua2017efficient} and achieve desirable performance in noiseless environments and Gaussian noise with small variance. However, when the data is contaminated with large outliers, the performance of these traditional algorithms is unsatisfactory in general. This spurred further research on robust matrix and tensor completion, which is the main focus of our work.}

{

\noindent\textbf{Robust matrix and tensor completion.}
In robust matrix and tensor completion, the goal is to recover the low-rank matrix or tensor from corrupted partial observations. Following the method of robust principal component analysis (RPCA) \cite{candes2011robust,fan2019exactly}, a matrix or tensor can be completed by decomposing it into the sum of low-rank and sparse components. The low-rank component represents the actual noise-free low-rank matrix or tensor and the sparse component models the sparse outliers. }

{
The $\ell_1$-norm is widely utilized to constrain the sparse component \cite{chen2012low,kang2019robust}, and a number of $\ell_1$-norm-based robust completion algorithms have been proposed for different decomposition (factorization) models such as matrix factorization \cite{cambier2016robust}, Tucker \cite{goldfarb2014robust}, tensor ring  \cite{huang2020robust} and t-SVD \cite{jiang2019robust}. 
Other algorithms impose more flexible $\ell_p$-norm and $\ell_{p,\epsilon}$-norm constraints on the sparse component instead of the $\ell_1$-norm \cite{liu2021simulated,li2021robust}. 
}

{
The success of the aforementioned existing robust methods is largely dependent on the assumption that the outliers are sparse -- otherwise, their performance may greatly degrade. 
In this paper, we develop a new two-stage framework with a new robust tensor completion algorithm to improve the performance under heavy noise and data corruption. } 

{
\noindent\textbf{Patch-based matrix and tensor completion.}
To further improve the completion performance on visual data, the patch-based method has been recently introduced to matrix and tensor completion. Similar to patch-based image and video denoising methods \cite{dabov2007image,maggioni2012nonlocal}, these methods apply block-matching 3D (BM3D) \cite{dabov2007image} or BM4D \cite{maggioni2012nonlocal} to find similar patches across the spatial domain. Then, the completion is applied to matrices or tensors constructed from similar patches. 
A variety of patch-based completion algorithms incorporating different matrix and tensor completion methods have been developed, such as \cite{li2015non,zhang2019nonlocal,ding2021tensor}. Also, in \cite{zhao2022tensor}, a tensor completion algorithm is proposed utilizing local and non-local patch completion as the regularization terms of the global tensor completion. 
}

{
Similar to the traditional completion algorithms, current patch-based methods also suffer from performance degradation in the presence of outliers. One idea would be to replace the completion methods in a path-based framework with robust ones. However, 
directly applying block-matching to robust tensor completion tasks may result in inaccurate matching results due to missing entries and outliers, which in turn affects the completion performance. 
In this paper, we propose an efficient method called `patch jitter' to directly bypass the block-matching procedure. Further, the global completion result is incorporated into local patch refinement to further improve performance. 
}

\section{PRELIMINARY}
\label{sec:background}

\textbf{Notation:} {In this paper, we adopt tensor notation similar to \cite{yu2019tensor,huang2020provable}.
Uppercase calligraphic letters are used to denote tensors (e.g., ${\mathcal{X}}$), uppercase boldface letters for matrices (e.g., ${\mathbf{X}}$), lowercase boldface letters for vectors (e.g., ${\mathbf{x}}$) and lowercase letters for scalars  (e.g., ${x}$).} {An} $N$-order tensor is defined as ${{\mathcal{X}}}\in{\mathbb{R}}^{I_1\times \dots \times I_N}$, where $I_i, i=1,\ldots, N$ is the dimension of the $i$-th way of the tensor. ${\mathcal{X}}_{i_1\ldots i_N}$ denotes the $(i_1,i_2,...,i_N)$-th entry of tensor ${\mathcal{{X}}}$, and ${\mathbf{X}}_{i,j}$ the $(i,j)$-th entry of matrix ${\mathbf{{X}}}$. The Frobenius norm of a tensor is defined as $\|{{\mathcal{X}}}\|_F=\sqrt{\sum_{i_1\ldots i_N}|{{\mathcal{X}}_{i_1\ldots i_N}}|^2}$. The product $\mathcal{A}\circ\mathcal{B}$ denotes the Hadamard (element-wise) product of tensors $\mathcal{A}$ and $\mathcal{B}$. For a scalar $n, [n]:=\{1,\ldots, n\}$.

\subsection{Tensor ring model}
We briefly review the definition of tensor ring decomposition.
\begin{mydef}
{(\cite[Section 2]{zhao2016tensor} \textbf{TR decomposition})} Given TR rank $[r_1,\ldots,r_N]$, in tensor ring (TR) decomposition a high-order tensor $\mathcal{X} \in \mathbb{R}^{I_{1} \times \cdots \times I_{N}}$ is represented as a sequence of circularly contracted 3-order core tensors $\mathcal{U}_k\in\mathbb{R}^{r_k\times I_k\times r_{k+1}},k=1,\ldots,N$, with $r_{N+1}=r_1$. Specifically, the element-wise relation of tensor $\mathcal{X}$ and its TR core tensors $\{\mathcal{U}_k\}_{k=1}^N$ is defined as
\[
\mathcal{X}_{i_{1} \ldots i_{N}}=\operatorname{Tr}\left(\prod_{k=1}^N\mathcal{U}_k(:,{{i_k}},:)\right)\:,
\]
where $\mathcal{U}_k(:,{{i_k}},:)\in\mathbb{R}^{r_k\times r_{k+1}}$ is the ${{i_k}}$-th slice matrix of $~\mathcal{U}_k$ along mode-$2$, and $\operatorname{Tr}(\cdot)$ is the matrix trace operator.
\label{def:TRdecomp}
\end{mydef}
%
Based on the definition above, the authors in \cite{yu2019tensor,huang2020provable} proposed a new circular TR unfolding scheme, {defined below.}
\begin{mydef}{(\cite[Section 2.2]{huang2020provable} \textbf{Tensor ring unfolding})}
Given an N-order tensor $\mathcal{X} \in \mathbb{R}^{I_{1} \times \cdots \times I_{N}}$, its tensor ring (TR) unfolding $\mathbf{X}_{\langle k, d\rangle} \in$ $\mathbb{R}^{\prod_{i=k}^{k+d-1} I_{i} \times \prod_{j=k+d}^{k+N-1} I_{j}}${, $k,d\in \{1,\ldots,N$\},} is a matrix whose entries are defined through the relation $(\mathbf{X}_{\langle k, d\rangle })_{s,t}=\mathcal{X}_{i_1\ldots i_N}$ with
$$
s\!=\!1+\!\!\sum_{c=k}^{k+d-1}\!\left(i_{c}-1\right) \prod_{j=k}^{c-1} I_{j}~,~~
t\!=\!1+\!\!\!\sum_{c=k+d}^{k+N-1}\!\left(i_{c}-1\right) \!\!\prod_{j=k+d}^{c-1}\!\!I_{j}
$$
where $I_{k+N}=I_k, i_{k+N}=i_k$ for {$1\leq k \leq N$}. In practice, $\mathbf{X}_{\langle k, d\rangle}$ can be generated by first permuting $\mathcal{X}$ with order $[k, \ldots, N, 1, \ldots, k-1]$, then performing unfolding along the first $d$ modes.
\label{def:TRunfolding}
\end{mydef}
\begin{mytheo} 
{(\cite[Section 2]{yu2019tensor})}
Assume $\mathcal{X} \in \mathbb{R}^{I_{1} \times \cdots \times I_{N}}$ is $N$th-order tensor with TR rank $[r_1, r_2, . . . , r_N]$, then for each unfolding matrix $\mathbf{X}_{\langle k,d\rangle}$
\begin{equation}
\operatorname{rank}(\mathbf{X}_{\langle k,d\rangle})\leq r_kr_{k+d} 
\label{eq:xkdrank}
\end{equation}
with $r_{i+N}=r_i, i=1,\ldots, N$, where { $\operatorname{rank}(\mathbf{X})$ denotes the rank of matrix $\mathbf{X}$.}
\end{mytheo}

\begin{figure*}[tb]
\centering
\includegraphics[width=0.98\linewidth]{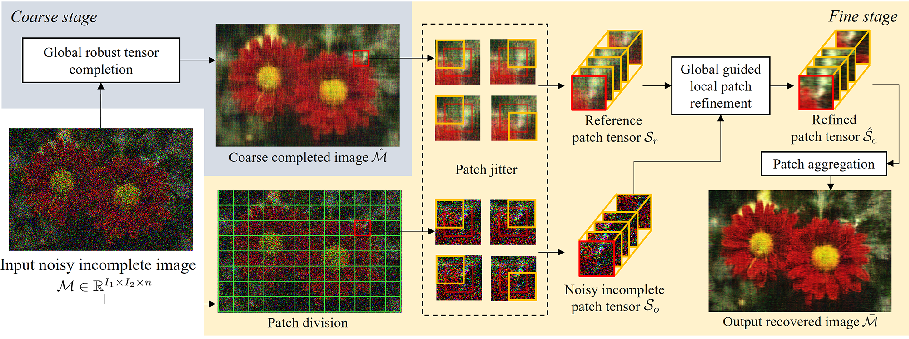}
\caption{Proposed two-stage coarse-to-fine robust tensor completion framework for visual data. In the coarse stage (blue region), the robust tensor completion algorithm is applied to the entire image. In the fine stage (yellow region), for each patch in the divided image, local patch refinement guided by the global completion result is applied to a corresponding patch tensor obtained from {the} patch jitter.}
\label{fig:PHQTRC}
\end{figure*}

\subsection{Tensor ring completion}
Given {an} $N$-order tensor $\mathcal{M}\in{\mathbb{R}}^{I_1\times \dots\times I_N}$, and {index set $\Omega\subseteq{[I_1]\times\ldots\times[I_N]}$}, tensor ring completion is the problem of filling in the missing entries of tensor $\mathcal{M}$ using the observed entries indexed by set $\Omega$ and the {low-rank} property. This problem can be formulated as
\begin{equation}
\min_{\mathcal{X}}\operatorname{rank}_{tr}(\mathcal{X})\text{, s.t. } \mathcal{P}\circ\mathcal{X}=\mathcal{P}\circ\mathcal{M}
\label{eq:tc_st}
\end{equation}
where the mask tensor $\mathcal{P}\in{\mathbb{R}}^{I_1\times \dots\times I_N}$ is set as
\begin{equation}
{\cal{P}}_{i_1\ldots i_N}=\left\{\begin{array}{cl}
1, & (i_1,\ldots,i_N) \in \Omega \\
0, & \text {otherwise}
\end{array}\right.
\label{eq:mask}
\end{equation}
and $\operatorname{rank}_{tr}(\mathcal{X})$ denotes the tensor ring rank of tensor $\mathcal{X}$. 
According to \eqref{eq:xkdrank}, one can further solve the tensor ring completion task using the following optimization problem \cite{huang2020provable,huang2020robust}:
\begin{equation}
\min_{\mathcal{X}}\sum_{k=1}^N \beta_{k}\operatorname{rank}(\mathbf{X}_{\langle k,d\rangle}) ,\text{~s.t.~} \mathcal{P}\circ\mathcal{X}=\mathcal{P}\circ\mathcal{M}
\label{eq:trc}
\end{equation}
where $\{\beta_k\}_{k=1}^N$ are weight parameters.

\section{Two-stage Coarse-to-fine robust tensor completion framework}
\label{sec:framework}

Our goal is to perform robust tensor completion of visual data with a large number of outliers. To this end, we develop a two-stage coarse-to-fine tensor completion framework, illustrated in Fig. \ref{fig:PHQTRC}. Given a noisy, partially observed image tensor $\mathcal{M}\in\mathbb{R}^{I_1\times I_2\times n}$ ($n$ is $1$ and $3$ for gray and color images, respectively), in the first (global) stage, a robust tensor ring completion algorithm is applied to the entire tensor, yielding a coarse completion result. In the second (local) stage, we first divide the tensor into overlapping patches of size $m\times m\times n$ with overlap pixels $o$. Then, with the guidance of the global completion result, we perform local patch-based robust tensor ring refinement on each patch tensor constructed using a patch jitter. The final completion result is obtained by aggregating the refined local patches. In the following, we describe each stage. The details of the robust recovery algorithms are discussed in Section \ref{sec:HQTRC}. 

\subsection{Global robust tensor completion with M-estimator and tensor ring rank}

In robust tensor completion, the predominant measure of error is the $\ell_1$-norm of the error residual \cite{goldfarb2014robust,huang2020robust,jiang2019robust}. The $\ell_1$-norm-based tensor completion algorithms aim to solve 
\begin{equation}
\min_{\mathcal{X}}\operatorname{rank}_t(\mathcal{X})+\lambda\|\mathcal{E}\|_1 ,\text{~s.t.~} \mathcal{P}\circ(\mathcal{X}+\mathcal{E})=\mathcal{P}\circ\mathcal{M}\:,
\label{eq:rtrcl1}
\end{equation}
where $\operatorname{rank}_t(\mathcal{X})$ denotes the rank of tensor $\mathcal{X}$, which varies depending on different definitions of the tensor rank. For the optimization problem \eqref{eq:rtrcl1}, it is always assumed that the error term $\mathcal{E}$ is sparse, i.e., {there} are only {a} few outliers. In the case where a large number of observed entries are perturbed by outliers, $\mathcal{E}$ is no longer sparse in general, and a solution to \eqref{eq:rtrcl1} is unreliable.

\begin{figure}[htbp]
\centering
\includegraphics[width=0.98\linewidth]{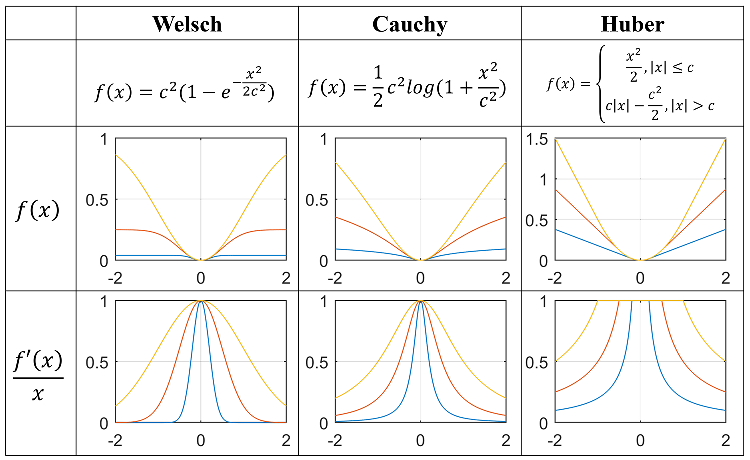}
\caption{Illustration of loss functions of M-estimators (top) and corresponding weight function (bottom) with different shape parameter $c$ (Blue: $c=0.2$, red: $c=0.5$, yellow: $c=1$).}
\label{fig:estimator}
\end{figure}

The M-estimator has been widely used in robust statistics due to its robustness to outliers. Given a tensor $\mathcal{X}$, its M-estimator $F(\mathcal{X})$ can be formed as a sum of functions of the data, i.e., $F(\mathcal{X})=\sum_{i_1\ldots i_N}f(\mathcal{X}_{i_1\ldots i_N})$, where $f(.)$ is a function with certain properties \cite{tyler1987distribution}. Compared with the $\ell_1$-norm, the M-estimator is {differentiable} at $0$, and is more flexible with different choices of a shape parameter (see Fig. \ref{fig:estimator}). In this work, we introduce an M-estimator with adaptive parameter selection to the robust tensor ring completion task.

By introducing the M-estimator in \eqref{eq:trc}, we obtain the unconstrained M-estimator-based robust tensor ring completion optimization problem
\begin{equation}
\min_{\mathcal{X}}\sum_{k=1}^N \beta_{k}\operatorname{rank}(\mathbf{X}_{\langle k,d\rangle})+{\lambda \sum_{i_1\ldots i_N}\mathcal{P}_{i_1\ldots i_N}f(\mathcal{E}_{i_1\ldots i_N})\:,}
\label{eq:rtrcm}
\end{equation}
{where $\mathcal{E}_{i_1\ldots i_N}=\mathcal{X}_{i_1\ldots i_N}-\mathcal{M}_{i_1\ldots i_N}$.}

In our work, we use three functions for M-estimators: Huber function, Welsch function \cite{dennis1978techniques} and Cauchy function shown in Fig. \ref{fig:estimator}. The Welsch and Cauchy functions yield a type of redescending M-estimators, which also satisfy that $\lim _{x \rightarrow+\infty}f^{\prime}(x)=0$. In \cite{yang2015robust}, a redescending M-estimator is introduced for low-Tucker-rank tensor completion, solved using a block coordinate descent method. However, the Tucker-rank-based method is not applicable to tensor ring-based methods due to the difference in the rank model. Also, its performance is limited by the low convergence speed of {the} gradient-based method. In the next section, we will develop a more general and efficient solution using a half-quadratic method for the tensor ring model.

The global completion can identify most of the reliable observed entries, i.e., the clean unperturbed observed entries {and the observed entries with small perturbation}. However, the global completion performance may still be limited due to insufficient reliable information for completion or disturbance by a small number of unrecognized noisy entries. On the other hand, patch-based methods can yield better completion performance than global ones by performing completion on similar patches \cite{zhang2019nonlocal,ding2021tensor}. To further improve the completion performance, we propose a new refinement process on local patches which incorporates both global and local information. In the following parts, we will introduce our proposed local-based method.

\subsection{Local patch tensor construction using patch jitter}
Patch-based methods have been widely used in visual data processing \cite{dabov2007image,maggioni2012nonlocal}. In tensor completion, a patch tensor is created using block-matching across the spatial domain \cite{xie2018tensor,zhang2019nonlocal,ding2021tensor}. Existing block-matching methods presume that the data entries are not perturbed by outliers, such that similar patches can be accurately matched. Similar patches are then stacked to a patch tensor, on which completion can be applied. However, in our setting, the distance between patches will be biased due to the existence of outliers, which would deteriorate the results of block-matching. 

By contrast, instead of using block-matching to find similar patches, in this work we directly apply patch jitter on each patch to generate a patch tensor. Specifically, given a patch of size $m\times m\times n$, we generate its neighboring patches with jitter length $l$, i.e., the $(2l+1)^2$ number of patches in a window of size $(2l+1)\times(2l+1)$ centered at the original patch. Note that the original patch is also included in the neighboring patch set. Then, the $(2l+1)^2$ patches are stacked in a patch tensor $\mathcal{S}\in\mathbb{R}^{m\times m\times n\times (2l+1)^2}$. To match the patches at the corners and boundaries of the frames, the tensor $\mathcal{M}$ is first padded by mirroring $l$ pixels at all boundaries and corners, resulting in a padded tensor {${\mathcal{M}}_p$} of size ${(I_1+2l)\times (I_2+2l)\times n}$.

\begin{figure}[tb]
\centering
\includegraphics[width=0.98\linewidth]{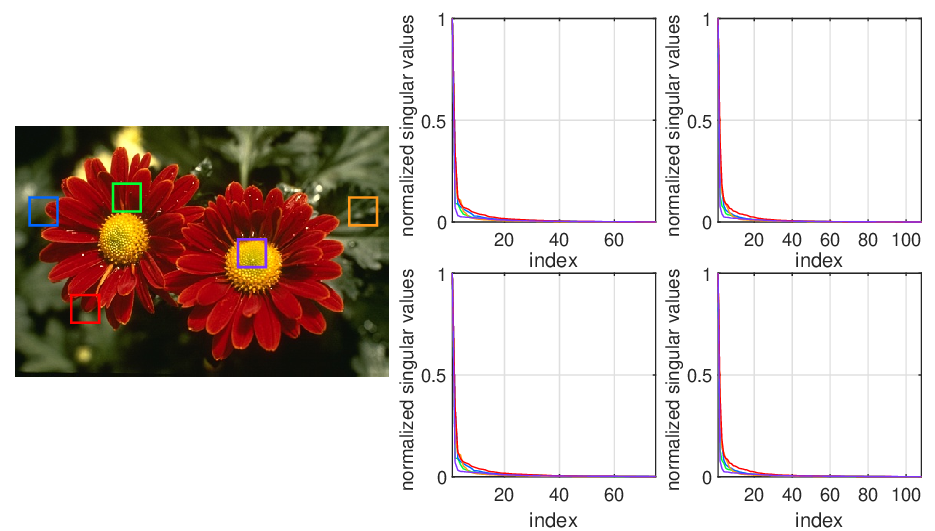}
\caption{Left: image `flower' from Berkeley Segmentation Dataset \cite{MartinFTM01}, rectangles: $5$ selected patches with size $36\times36\times3$. Right: normalized {singular values} of TR unfolding matrices of the $5$ patch tensors obtained using patch jitter with $l=2$. Top-left: $k=1,d=2$, top-right: $k=2,d=2$, bottom-left: $k=3,d=2$, bottom-right: $k=4,d=2$.}
\label{fig:insight}
\end{figure}

We briefly give insight into the patch jitter procedure. An example of {a} patch jitter with a fully observed image without outliers is shown in Fig. \ref{fig:insight}. We pick $5$ patches at different locations (marked by green rectangles). For each patch, the patch tensor is created using patch jitter with $l=2$. The normalized {singular values} (w.r.t. maximum {singular value}) of TR unfolding matrices of the patch tensors are shown in the right of Fig. \ref{fig:insight}. As can be seen, with a small offset around the original patch, the obtained patch tensor can be well approximated by a low tensor ring rank tensor. Therefore, for a partially observed image, the patch tensor generated using {a} patch jitter can be well completed using a tensor ring completion algorithm. Moreover, compared with traditional patch-based methods, the jitter operation does not require block-matching, thereby avoiding incorrect matching due to outliers and greatly reducing the computational cost.  

\subsection{Global completion guided local patch tensor refinement}
After constructing patch tensors using {the} patch jitter, we apply a local patch refinement process {to} each patch tensor. We utilize the global coarse completion result to help identify the outliers in the patch tensor, as well as give a good initialization to the missing entries of the patch tensor. 

{Assume we have obtained the (coarsely) completed tensor (denoted $\hat{\mathcal{M}}$) from the tensor $\mathcal{M}$ using \eqref{eq:rtrcm}.} Given a patch tensor {$\mathcal{S}_o\in\mathbb{R}^{m\times m \times n \times (2l+1)^2}$ from $\mathcal{M}_p$, we extract the patches from the same locations in the padded completed tensor $\hat{\mathcal{M}}_p$} and stack them to a reference patch tensor $\mathcal{S}_r$. Then, the missing entries in {$\mathcal{S}_o$} are filled with corresponding entries in $\mathcal{S}_r$, resulting in a combined tensor $\mathcal{S}_c$ with entries
{
\begin{equation}
{(\mathcal{S}_c)}_{i_1\ldots i_N}\!=\!\left\{\begin{array}{ll}
{(\mathcal{S}_o)}_{i_1\ldots i_N},&\!(i_1,\ldots,i_N) \!\in\! \Omega_s \\
{(\mathcal{S}_r)}_{i_1\ldots i_N}, & \text {otherwise }
\end{array}\right.
\label{eq:weight}
\end{equation}
where $\Omega_s$ denotes the observation index set of $\mathcal{S}_o$.}


In order to represent the different confidence levels of each entry, we use a soft weighting strategy \cite{yu2014click} in which we assign different weights to each entry of the combined patch tensor $\mathcal{S}_c$. In particular, defining a weight tensor $\mathcal{W}$ with the same size {as} $\mathcal{S}_c$, each element of $\mathcal{W}$ is obtained as
\begin{equation}
\mathcal{W}_{i_1\ldots i_N}\!=\!\left\{\begin{array}{ll}
\!\!\exp\!\left(\!-\frac{({(\mathcal{S}_c)}_{i_1\ldots i_N}\!-\!{(\mathcal{S}_r)}_{i_1\ldots i_N})^2}{2{\sigma}^2}\right)\!,&\!\!(i_1,\ldots,i_N) \!\in\! \Omega_s \\
\!\!{v},&\!\!\text{otherwise }
\end{array}\right.
\label{eq:weight}
\end{equation}
where {$\sigma$} is a parameter controlling the similarity. Specifically, for entries of $\mathcal{S}_c$ that are originally observed (i.e., {$(i_1,\ldots,i_N)\in\Omega_s$}), the weight is assigned in terms of its distance from the corresponding entry in the reference patch tensor $\mathcal{S}_r$. For the entries of $\mathcal{S}_c$ that were filled from $\mathcal{S}_r$, the weights {are all} set to some value {$v$}.

{Inspired by adaptive parameter selection for the M-estimator which will be presented in Section \ref{subsec:adaptive}, we adaptively determine $\sigma_{\min}$ and $v_{\max}$ using 
\begin{equation}
\begin{aligned}
\sigma&=\max\left\{\eta_{\sigma}\left(\max\{({\mathbf{d}_{{\Omega}_s}})_{(0.25)},({\mathbf{d}_{{\Omega}_s}})_{(0.75)}\}\right)^{-1},\sigma_{\min}\right\}\\
v&=\min\left\{\eta_{v}\max\{({\mathbf{d}_{{\Omega}_s}})_{(0.25)},({\mathbf{d}_{{\Omega}_s}})_{(0.75)}\},v_{\max}\right\}
\end{aligned}
\label{eq:sigmav}
\end{equation}
where $\mathbf{d}_{{\Omega}_s}\in\mathbb{R}^{|\Omega_s|\times 1}$ denotes the vector composed of entries $\mathcal{D}_{i_1\ldots i_N}\!=\!{(\mathcal{S}_c)}_{i_1\ldots i_N}-{(\mathcal{S}_r)}_{i_1\ldots i_N},(i_1,\ldots,i_N)\!\in\!\Omega_s$, $\mathbf{y}_{(q)}$ denotes the $q$-th quantile of $\mathbf{y}$, $\eta_{\sigma}$ and $\eta_{v}$ are free parameters to be chosen, $\sigma_{\min}$ is a lower bound on $\sigma$, and $v_{\max}$ is an upper bound on $v$. }

We can readily formulate the local patch refinement as a weighted robust tensor recovery problem
\begin{equation}
\min_{\mathcal{X}}\sum_{k=1}^N \beta_{k}\operatorname{rank}(\mathbf{X}_{\langle k,d\rangle})+\lambda \sum_{i_1\ldots i_N}\mathcal{W}_{i_1\ldots i_N}f((\mathcal{E}_c)_{i_1\ldots i_N})\:,
\label{eq:wtrm}
\end{equation}
where $(\mathcal{E}_c)_{i_1\ldots i_N}=\mathcal{X}_{i_1\ldots i_N}-{(\mathcal{S}_c)}_{i_1\ldots i_N}$.

Note that \eqref{eq:wtrm} can be obtained by replacing the binary indicator tensor $\mathcal{P}$ in \eqref{eq:rtrcm} with a weight tensor $\mathcal{W}$ with entries from $[0,1]$. Therefore, \eqref{eq:rtrcm} can be viewed as a special case of \eqref{eq:wtrm} with binary weights. In Section \ref{sec:HQTRC}, we propose a half quadratic-based algorithm that can solve both \eqref{eq:rtrcm} and \eqref{eq:wtrm}.

The completion and refinement processes of the proposed framework are summarized in Algorithm \ref{alg:online}. We remark that the framework can also be extended to video data, where an additional temporal dimension is added. In this case, for a video with $f$ frames, the patch will be of size $m\times m \times n \times f$ and the corresponding patch tensor is a $5$th-order tensor of size $m\times m \times n \times f \times (2l+1)^2$.

\begin{algorithm}
\caption{Coarse-to-fine robust tensor ring completion (C2FRTRC)}
\begin{algorithmic}[1]
 \REQUIRE Partially observed image tensor $\mathcal{M}\in\mathbb{R}^{I_1\times I_2\times n}$, patch generation parameters $m$, $o$, patch jitter parameter $l$, {parameters $\eta_{\sigma}$, $\eta_{v}$, $\sigma_{\min}$ and $v_{\max}$}.
 \STATE Complete $\mathcal{M}$ using \eqref{eq:rtrcm} and obtain the completed tensor $\hat{\mathcal{M}}$.
 \STATE Pad the tensors $\mathcal{M}$ and $\hat{\mathcal{M}}$ to {$\mathcal{M}_p$ and $\hat{\mathcal{M}}_p$}.
 \STATE Divide $\mathcal{M}_p$ into patches according to size $m$ and overlap pixels $o$ and create patch set $\mathcal{I}$.
 \FOR {each patch in $\mathcal{I}$}
 \STATE Construct patch tensor $\mathcal{S}_o$ using $(2l+1)^2$ neighbor patches around the location in {$\mathcal{M}_p$}. 
 \STATE Construct reference patch tensor ${(\mathcal{S}_r)}$ using $(2l+1)^2$ neighbor patches around the location in {$\hat{\mathcal{M}}_p$}.
 \STATE {Construct combined tensor $\mathcal{S}_c$ using $\mathcal{S}_o$ and $\mathcal{S}_r$, and obtain refined patch tensor $\hat{\mathcal{S}_c}$ by solving \eqref{eq:wtrm}.}
 \ENDFOR
 \STATE Obtain refined completed image tensor $\bar{\mathcal{M}}$ by aggregating patches from all patch tensors {$\hat{\mathcal{S}_c}$} according to $\mathcal{I}$ and {removing} padded border pixels.
\ENSURE Completed image tensor $\bar{\mathcal{M}}$.
\end{algorithmic}
\label{alg:online}
\end{algorithm}

\section{Half-quadratic approach to weighted robust tensor recovery}
\label{sec:HQTRC}

In this section, we aim to solve the following optimization problem, which combines both \eqref{eq:rtrcm} and \eqref{eq:wtrm}:
\begin{equation}
\min_{\mathcal{X}}\Phi(\mathcal{X})+\lambda\!\!\!\sum_{i_1\ldots i_N}\!\mathcal{W}_{i_1\ldots i_N}f(\mathcal{E}_{i_1\ldots i_N})
\label{eq:rtrcm2}
\end{equation}
where $\Phi(\mathcal{X})=\sum_{k=1}^N \beta_{k}\operatorname{rank}(\mathbf{X}_{\langle k,d\rangle})$, and the entries of $\mathcal{W}_{i_1\ldots i_N}$ are in the range $[0,1]$ with $\mathcal{W}_{i_1\ldots i_N}=0,(i_1,\ldots,i_N) \notin \Omega$. We develop {an} HQ-based approach to efficiently solve \eqref{eq:rtrcm2}. We also propose an adaptive parameter selection strategy and discuss the property of the HQ-based solution. 

\subsection{Half-quadratic approach to the non-convex program} 
We present a half-quadratic (HQ) method to solve the M-estimator-based optimization problem in \eqref{eq:rtrcm2}. HQ methods have been broadly applied in non-quadratic optimization \cite{he2014robust}. Instead of directly optimizing a complex non-quadratic objective,
HQ transforms the non-quadratic loss function to a half-quadratic one. Specifically, there exists a strictly convex and decreasing dual function $\varphi(.)$ such that
minimizing the loss function $f(t)$ with respect to (w.r.t.) $t$ is equivalent to minimizing an augmented cost function in an enlarged parameter space $\{t,q\}$, i.e., \cite{charbonnier1997deterministic,he2014robust}
\begin{equation}
\label{eq:HQ2}
\min_{t} f(t)=\min_{t,q}\frac{1}{2}q t^{2}+\varphi(q)\:.
\end{equation}
Therefore, by substituting \eqref{eq:HQ2} in the M-estimator, the minimization of function $\sum_{i_1\ldots i_N}\mathcal{W}_{i_1\ldots i_N}f(\mathcal{E}_{i_1\ldots i_N})$ becomes
\begin{equation}
\begin{aligned}
&\min_{\mathcal{X}}\!\sum_{i_1\ldots i_N}\mathcal{W}_{i_1\ldots i_N}f(\mathcal{E}_{i_1\ldots i_N})\\
&\!=\min_{\mathcal{X},\mathcal{Q}}\sum_{i_1\ldots i_N}\!\!\left(\frac{1}{2}\mathcal{W}_{i_1\ldots i_N}\mathcal{Q}_{i_1\ldots i_N}\mathcal{E}_{i_1\ldots i_N}^2\!+\!\mathcal{W}_{i_1\ldots i_N}\varphi({\cal{Q}}_{i_1\ldots i_N})\!\right)
\end{aligned}
\label{eq:hq_w2}
\end{equation}
Hence, \eqref{eq:wtrm} {is equivalent to the following problem}
\begin{equation}
\begin{aligned}
\min_{\mathcal{X},\mathcal{Q}}\Phi(\mathcal{X})\!+\!\frac{\lambda}{2}\|\sqrt{\mathcal{W}}\circ\sqrt{\mathcal{Q}}\circ(\mathcal{X}-\mathcal{M})\|_F^2+\lambda\Psi_\mathcal{W}(\mathcal{Q})
\end{aligned}
\label{eq:tc_hq}
\end{equation}
where $\Psi_{\mathcal{W}} \left( {{\cal{Q}}}\right)=\sum\nolimits_{i_1\ldots i_N}\mathcal{W}_{i_1\ldots i_N}\varphi \left( {{{\cal{Q}}_{i_1\ldots i_N}}} \right)$. 

The problem above is a reweighted tensor ring completion problem, and one could use alternating minimization to solve it. Specifically, by fixing tensor $\mathcal{X}$, tensor $\mathcal{Q}$ can be found by solving 
\eqref{eq:tc_hq} with fixed residual $\mathcal{E}$. According to \cite[Theorem 1]{charbonnier1997deterministic}, the optimal solution ${q}^*$ in the RHS of \eqref{eq:HQ2} can be obtained as ${q}^*=\frac{f'(t)}{t}$. Thus, we obtain each entry $\mathcal{Q}_{i_1\ldots i_N}$ as
\begin{equation}
\label{eq:HQW}
\mathcal{Q}_{i_1\ldots i_N}=\frac{f'(\mathcal{E}_{i_1\ldots i_N})}{\mathcal{E}_{i_1\ldots i_N}}\:.
\end{equation}
Subsequently, given a fixed ${\mathcal{Q}}$, \eqref{eq:tc_hq} becomes the double-weighted tensor completion problem
\begin{equation}
\label{eq:WTF}
\min_{\mathcal{X}}\Phi(\mathcal{X})+\frac{\lambda}{2}\|\sqrt{\mathcal{W}}\circ\sqrt{\mathcal{Q}}\circ(\mathcal{X}-\mathcal{M})\|_F^2\:.
\end{equation}
The weighting tensor ${\mathcal{Q}}$ assigns different weights to each observed entry based on the error residual tensor $\mathcal{E}$. Fig. \ref{fig:estimator} depicts the weights in terms of the error, $x$, for different loss functions. One can observe that given a proper shape parameter $c$, a large error may lead to a small weight, so that the error statistics will not be unduly affected by large outliers. Specifically, when $c\rightarrow+\infty$, all entries of ${\mathcal{Q}}$ will be equal to 1 and \eqref{eq:tc_hq} reduces to a traditional second-order statistics-based completion method. In this case, the algorithm cannot alleviate the effect of outliers since all error residuals are treated equally.

\subsection{Adaptive parameter selection for M-estimator}
\label{subsec:adaptive}
Most M-estimators such as Huber, Cauchy and Welsch have a parameter $c$ to control the shape of the loss. Per the previous discussion, the weights $\mathcal{Q}$ based on the error residual play an important role in recognizing the outliers. As Fig. \ref{fig:estimator} shows, a relatively smaller $c$ can better reduce the effect of outliers. However, in practice convergence will be slower if $c$ is set to a small fixed value. Therefore, to improve both efficiency and accuracy, we use an adaptive kernel width selection method for the M-estimator. Specifically, the shape parameter is determined by
\begin{equation}
\label{eq:sigma}
{c=\max\left\{\eta_{c}\max\{({\mathbf{e}_{\Omega}})_{(0.25)},({\mathbf{e}_{\Omega}})_{(0.75)}\},c_{\min}\right\}}
\end{equation}
where $\mathbf{e}_{\Omega}\in\mathbb{R}^{|\Omega|\times 1}$ denotes the vector composed of entries $\mathcal{E}_{i_1\ldots i_N},i_1\ldots i_N\in\Omega$. The parameter $\eta_{c}$ controls the range of outliers, and $c_{\min}$ is a lower bound on $c$. Using the adaptive method above, $c$ is set to a relatively large value in the beginning to speed up convergence. As the {convergence} speed reduces, $c$ is decreased correspondingly and the effect of the outliers is gradually reduced.

\subsection{Truncated SVD-based algorithm}
To solve \eqref{eq:tc_hq}, we define the indicator function for $\mathbf{X}_{\langle k,d\rangle}, k=1,\ldots,N$ as \cite{yang2015robust}
\begin{equation}
\delta(\mathbf{X}_{\langle k,d\rangle})=\left\{\begin{array}{ll}
0, & \text { if } \operatorname{rank}(\mathbf{X}_{\langle k,d\rangle})\leq {{{r}_{kd}}} \\
+\infty, & \text { otherwise }
\end{array}\right.\:,
\end{equation}
where {${{r}_{kd}}=r_kr_{k+d}$}. Thus, the minimization is expressed as
\begin{equation}
\min _{\mathcal{X},\mathcal{Q}} \sum_{k=1}^{N}\beta_k\delta(\mathbf{X}_{\langle k,d\rangle})+\frac{\lambda}{2}\|\sqrt{\mathcal{W}}\circ\sqrt{\mathcal{Q}}\circ(\mathcal{M}-\mathcal{X})\|_{F}^2+\lambda\Psi(\mathcal{Q})
\label{eq:trnn_o}
\end{equation}
We devise an {alternating direction method of multipliers (ADMM)} method to solve \eqref{eq:trnn_o}. In particular, we introduce the dual variables $\{\mathcal{Z}^{(k)}\}_{k=1}^N$ and rewrite \eqref{eq:trnn_o} as
\begin{equation}
\begin{aligned}
\min _{\mathcal{X},\mathcal{Q},\mathcal{Z}^{(k)}} &\sum_{k=1}^{N}\beta_k\delta(\mathbf{Z}_{\langle k,d\rangle}^{(k)})\!+\!\frac{\lambda}{2}\|\sqrt{\mathcal{W}}\!\circ\!\sqrt{\mathcal{Q}}\!\circ\!(\mathcal{M}\!-\!\mathcal{X})\|_{F}^2\\
&+\!\lambda\Psi(\mathcal{Q})\text{~~~s.t.~~} \mathcal{Z}^{(k)}\!=\!\mathcal{X}, ~k\!=\!1,\ldots,N
\end{aligned}
\end{equation}
The augmented Lagrangian function can be written as
\begin{equation}
\begin{aligned}
&{\mathscr{L}_{\mu}}(\mathcal{X},\mathcal{Q},\mathcal{Z}^{(1)},\ldots,\mathcal{Z}^{(N)},\mathcal{G}^{(1)},\ldots,\mathcal{G}^{(N)})\\
=\!&\sum_{k=1}^{N}\!\left(\beta_k\delta(\mathbf{Z}_{\langle k,d\rangle}^{(k)})\!+\!\langle\mathcal{G}^{(k)},\mathcal{Z}^{(k)}\!-\!\mathcal{X}\rangle\!+\!\frac{\mu}{2}\|\mathcal{Z}^{(k)}\!-\!\mathcal{X}\|_F^2\right)\\
&\!+\!\frac{\lambda}{2}\|\sqrt{\mathcal{W}}\circ\sqrt{\mathcal{Q}}\circ(\mathcal{M}-\mathcal{X})\|_{F}^2+\lambda\Psi(\mathcal{Q})
\end{aligned}
\end{equation}
where $\{\mathcal{G}^{(k)}\}_{k=1}^N$ are the dual variables and $\mu$ is the step size. One can alternatively update each variable while fixing the others:

1) Update $c$ and $\mathcal{Q}$: First, $c$ is estimated using \eqref{eq:sigma}. Then, for each element $\mathcal{Q}_{i_1\ldots i_N}$, the optimal solution can be directly obtained using \eqref{eq:HQW}.

2) Update $\mathcal{Z}^{(k)}$: For each $\mathcal{Z}^{(k)}, k=1,\ldots,N$, the optimal solution can be obtained by solving
\begin{equation}
\begin{aligned}
\mathcal{Z}^{(k)}\!\!
=\!\arg\min_{\mathcal{Z}}\left(\|\mathcal{Z}\!-\!(\mathcal{X}\!-\!\frac{1}{\mu}\mathcal{G}^{(k)})\|_F^2\right)\text{~s.t.} \operatorname{rank}(\mathbf{Z}_{\langle k,d\rangle}\!)\!\leq \!{{{r}_{kd}}}
\end{aligned}
\label{eq:trnn_z}
\end{equation}
This is a {low-rank} approximation problem which has {an} optimal solution \cite{eckart1936approximation}
\begin{equation}
\mathcal{Z}^{(k)}=\operatorname{fold_{\langle k,d\rangle}}\!\left(\Pi_{{{{r}_{kd}}}}\!(\mathbf{X}_{\langle k,d\rangle}-\frac{1}{\mu}\mathbf{G}_{\langle k,d\rangle}^{(k)})\right)\:,
\label{eq:trnn_z_cf}
\end{equation}
where $\Pi_r(.)$ is the truncated SVD (or hard thresholding) operator with rank $r$, and $\operatorname{fold_{\langle k,d\rangle}}(.)$ is the reverse operation of TR unfolding.

3) Update $\mathcal{X}$: Tensor $\mathcal{X}$ can be obtained as
\begin{equation}
\begin{aligned}
\mathcal{X}
=&\arg\min_{\mathcal{X}}\left(\frac{\lambda}{\mu}\|\sqrt{\mathcal{W}}\circ\sqrt{\mathcal{Q}}\circ(\mathcal{M}-\mathcal{X})\|_{F}^2\right.\\
&~~~~~~~~~~\left.+\sum_{k=1}^{N}\|\mathcal{X}-(\mathcal{Z}^{(k)}+\frac{1}{\mu}\mathcal{G}^{(k)})\|_F^2\right)\:.
\end{aligned}
\end{equation}
By taking the derivative of $\mathcal{X}$ and setting it to be the zero tensor, we obtain the optimal solution
\begin{equation}
\mathcal{X}=\mathcal{L}+\frac{\lambda\mathcal{W}\circ\mathcal{Q}}{\lambda\mathcal{W}\circ\mathcal{Q}+\mu N}\circ(\mathcal{M}-\mathcal{L})\:,
\label{eq:trnn_x}
\end{equation}
where {the division of tensors is computed element-wise and} $\mathcal{L}=\frac{1}{N}\sum_{k=1}^{N}(\mathcal{Z}^{(k)}+\frac{1}{\mu}\mathcal{G}^{(k)})$.

4) Update $\mathcal{G}^{(k)}$: For each $k$, $\mathcal{G}^{(k)}$ can be updated as
\begin{equation}
\mathcal{G}^{(k)}=\mathcal{G}^{(k)}+\mu(\mathcal{Z}^{(k)}-\mathcal{X})\:.
\label{eq:trnn_g}
\end{equation}
We name this algorithm Half-Quadratic-based Weighted Tensor Ring Recovery (HQWTRR), and its pseudocode is summarized in Algorithm 2. For global coarse completion, {$\{\mathcal{G}^{(k),0}\}_{k=1}^{N}$ and $\mathcal{X}^1$ are initialized as zero tensors,} and $\mathcal{W}$ is set according to \eqref{eq:mask}, i.e., entries $(i_1,\ldots,i_N) \in \Omega$ are set to 1 and 0 otherwise. While in the global-completion-guided local patch refinement, {for each patch $\mathcal{S}$, all entries of $\{\mathcal{G}^{(k),0}\}_{k=1}^{N}$ and $\mathcal{X}^1$ are initialized as the average value of entries of the corresponding reference patch $\mathcal{S}_r$,} and $\mathcal{M}$ is a fully observed tensor with $\mathcal{W}$ obtained from \eqref{eq:weight}.
\begin{algorithm}
\caption{HQWTRR for weighted robust tensor recovery}
\begin{algorithmic}[1]
 \REQUIRE Partially observed ${\mathcal{M}}$ with observation set $\Omega$, $\mathcal{W}$, $d$, $\mu$, $\alpha$, $\{{r}_{k}\}_{k=1}^N$, $\lambda$, {$\eta_{c}$} and $\epsilon$
 \STATE initial tensors $\{\mathcal{G}^{(k),0}\}_{k=1}^{N}$, $\mathcal{X}^1$, set $\mathcal{W}_{i_1\ldots i_N}=0$ for $(i_1,\ldots,i_N) \notin \Omega$, $t=1$\\
 \REPEAT
 \STATE estimate ${c}^{t+1}$ using \eqref{eq:sigma}.
 \STATE compute $\mathcal{Q}^{t+1}$ using \eqref{eq:HQW}.
 \STATE compute $\mathcal{Z}^{(k),{t+1}}$ for $k=1,\ldots, N$ using \eqref{eq:trnn_z_cf}.
 \STATE compute $\mathcal{X}^{t+1}$ using \eqref{eq:trnn_x}.
 \STATE compute $\mathcal{G}^{(k),{t+1}}$ for $k=1,\ldots, N$ using \eqref{eq:trnn_g}.
 \STATE update $\mu^{t+1}=\alpha\mu^{t}$
 \STATE $t=t+1$
 \UNTIL $|\|\mathcal{X}^{t-1}-\mathcal{X}^{t-2}\|_F/\|\mathcal{X}^{t-2}\|_F-\|\mathcal{X}^{t}-\mathcal{X}^{t-1}\|_F/\|\mathcal{X}^{t-1}\|_F|<\epsilon$
\ENSURE $\hat{\mathcal{M}}=\mathcal{X}^{t}$.
\end{algorithmic}
\label{alg:HQTRC}
\end{algorithm}
\subsection{Relation to prior tensor ring completion algorithms}
To better understand the relationship between the proposed algorithm and existing $\ell_2$-norm-based tensor ring completion algorithms, we first rewrite \eqref{eq:trnn_x}  element-wise as
\begin{equation}
\mathcal{X}_{i_1\ldots i_N}=\Theta_{i_1\ldots i_N}\mathcal{M}_{i_1\ldots i_N}\!+\!(1\!-\!\Theta_{i_1\ldots i_N})\mathcal{L}_{i_1\ldots i_N}
\label{eq:theta_hq}
\end{equation}
with $\Theta=\frac{\lambda\mathcal{W}\circ\mathcal{Q}}{\lambda\mathcal{W}\circ\mathcal{Q}+\mu N}$. When the regularization parameter $\lambda$ is set to a sufficiently large value compared with $\mu N$, \eqref{eq:theta_hq} reduces to
\begin{equation}
\begin{aligned}
\mathcal{X}=\left\{\begin{array}{cc}
    \mathcal{M}_{i_1\ldots i_N}, &  (i_1,\ldots,i_N) \in \Omega\\
    \mathcal{L}_{i_1\ldots i_N}, &  (i_1,\ldots,i_N) \notin \Omega
\end{array}
\right.
\end{aligned}
\label{eq:mq}
\end{equation}
Again, by replacing hard thresholding using {$\{{{r}_{kd}}\}_{k,d=1}^N$} in \eqref{eq:trnn_z} with a soft thresholding method, Algorithm 2 reduces to the traditional tensor ring nuclear norm minimization (TRNNM) method \cite{yu2019tensor} solving
\begin{equation}
\min_{\mathcal{X}}\sum_{k=1}^N \beta_{k}\|\mathbf{X}_{\langle k,d\rangle}\|_*+\lambda \|\mathcal{W}\circ(\mathcal{X}-\mathcal{M})\|_F^2\:.
\label{eq:trnn}
\end{equation}
Thus, TRNNM can be seen as a special case of HQWTRR.

When the regularization parameter $\lambda$ is properly chosen, the elements of $\mathcal{Q}$ will assign different weights to different values of the error residuals. It can be observed from Fig. \ref{fig:estimator} that a large error residual $\mathcal{E}_{i_1\ldots i_N}$ caused by an outlier may result in a small $\mathcal{Q}_{i_1\ldots i_N}$ (consequently a small ${\Theta}_{i_1\ldots i_N}$). In this case, the values of the entries with large error residuals will be dominated by the predicted value $\mathcal{Q}$ rather than $\mathcal{M}$. If the error residual is large enough, $\theta$ will be zero so the corresponding entry will be set to the corresponding entry in $\mathcal{Q}$, which amounts to treating it as a missing entry. In general, by assigning different weights to observed entries, the proposed algorithm can automatically identify the outliers.

\subsection{Convergence analysis}

The following theorem characterizes the convergence of the proposed algorithm. For simplicity, we define  $\mathcal{Z}_a=\{\mathcal{Z}^{(k)}\}_{k=1}^N$ and $\mathcal{G}_a=\{\mathcal{G}^{(k)}\}_{k=1}^N$.


\begin{mytheo}[HQWTRR convergence]
Let $\left\{(\mathcal{X}^t,\mathcal{Q}^t,\mathcal{Z}^t_a,\mathcal{G}^t_a\right)\}$ be a sequence generated by Algorithm \ref{alg:HQTRC} using the M-estimators defined in Fig. \ref{fig:estimator}. If {  $\|\mathcal{X}^{t+1}-\mathcal{X}^{t}\|_F^2<\infty$ for any $t\geq 1$,} and $\{\mathcal{G}^{(k),t}\}$ converges to some constant tensor $\mathcal{C}$ for all $k\!=\!1,\ldots,N$, then $\left\{\mathcal{X}^t\right\}$ will converge for an M-estimator parameter $c$ decreasing to $0$.
\label{thm:HQTRC}
\end{mytheo}

The proof is deferred to Appendix A. In the theorem, a sequence $\{c^t\}$ with $\lim_{t\rightarrow\infty}c^t=0$ is sufficient to ensure convergence of HQWTRR. In practice, adaptive parameter selection using \eqref{eq:sigma} yields a sequence $\{c^t\}$ that approaches a small $c_{\min}$, albeit not monotonically decreasing. Still, it yields desirable performance as shown in the experimental results. Since HQWTRR is a non-convex optimization problem due to the use of a truncated SVD, the convergence analysis of ADMM is very challenging in general without additional assumptions \cite{boyd2011distributed}. Hence, similar to \cite{jiang2014alternating,magnusson2015convergence}, the assumption of the convergence of $\{\mathcal{G}^{(k),t}\}$ are used in Theorem \ref{thm:HQTRC}. In practice, HQWTRR  using Algorithm \ref{alg:HQTRC} works very well without this assumption, which is verified in Section \ref{sec:results}.

{
\subsection{Complexity analysis}
We first analyze the complexity of the global coarse completion step. Given an $N$-order tensor $\mathcal{X}\in\mathbb{R}^{I_1\times I_2\times \dots \times I_N}$, for simplicity we assume the TR rank $r_1=\ldots=r_N=r$ and the tensor size $I_1=\ldots=I_N=I$. Then, the time complexity of updating $\{\mathcal{Z}^{(k)}\}_{k=1}^N$ in \eqref{eq:trnn_z_cf} using truncated SVD is $\mathcal{O}(I^Nr^2N)$. The update of $\mathcal{Q}$ incurs a complexity of $\mathcal{O}(I^N)$, and the complexity of updating $\{\mathcal{G}^{k}\}_{k=1}^N$ or $\mathcal{X}$ is $\mathcal{O}(I^NN)$. Therefore the total time complexity of the coarse stage is $\mathcal{O}(I^N(r^2+2)N)$.

For the local patch refinement step, assume the total number of patch tensors is $T_p$. The dimension of the tensor is $N_p$, the length of each dimension is $I_p$, and the elements of the TR rank are all set to $r_p$. Then, the total time complexity of the fine stage is $\mathcal{O}(I_p^{N_p}(r_p^2+2)T_p N_p)$. We should remark that tensor completion is independent for each patch tensor, hence parallel computation can be applied to further improve the computational efficiency. }

\section{Experimental results}
\label{sec:results}

We conduct experiments using both image and video data to verify the performance of the proposed algorithm. We compare with existing tensor completion algorithms using different tensor rank models, including {$\ell_1$-regularized sum of nuclear norm ($\ell_1$-SNN)\footnote{https://tonyzqin.wordpress.com/research} \cite{goldfarb2014robust}, Soft thresholding using Welsch loss (W-ST)\cite{yang2015robust}, tensor nuclear norm (TNN) \cite{zhang2016exact}, $\ell_1$-regularized TNN ($\ell_1$-TNN) \cite{jiang2019robust}, transformed nuclear norm-based total variation (TNTV)\footnote{https://github.com/xjzhang008/TNTV} \cite{qiu2021robust}, $\ell_{p}$-regularized tensor train completion ($\ell_{p}$-TTC)\footnote{https://github.com/LI-X-P/CodeofRobustTensorCompletion} \cite{liu2021simulated}, tensor ring nuclear norm (TRNN) \cite{yu2019tensor}, $\ell_1$ regularized TRNN ($\ell_1$-TRNN)\footnote{https://github.com/HuyanHuang/Robust-Low-rank-Tensor-Ring-Completion} \cite{huang2020robust} and $\ell_{p,\epsilon}$-regularized tensor ring completion ($\ell_{p,\epsilon}$-TRC) \cite{li2021robust}. All these algorithms are robust tensor completion algorithms except TRNN and TNN.} For the proposed algorithm, we use C2FRTRC to designate the two-stage algorithm described in Algorithm 1, which uses HQWTRR for both global completion and local refinement. For comparison, we also include the results of global completion alone (without the local refinement) obtained at the coarse stage using HQWTRR, {and local patch-only completion results without the global completion prior (i.e., setting $v_{\max}=0$ and $\sigma_{\min}=+\infty$). 
To distinguish these two single-stage methods from the two-stage C2FRTRC, in the experiments we name the global-only completion procedure `Half-Quadratic Tensor Ring Completion' (HQTRC), and the local patch-only completion algorithm `local patch-based robust tensor ring completion' (LPRTRC). For HQTRC, LPRTRC and C2FRTRC, we use the Cauchy loss function as the default.}

{Two visual data quality metrics are used, including peak signal-to-noise ratio (PSNR) and structural similarity (SSIM). For each experiment, the average PSNR/SSIM values are obtained over 20 Monte Carlo runs with different missing entries and noise realizations. For the proposed C2FRTRC framework in Algorithm 1, the patch size $m$ is set to $36$ and $20$ for image and video data, respectively. The pixel overlap $o$ is set to $o=\lceil m/5 \rceil$, and the jitter parameter $l$ is set to $2$. For HQTRC in Algorithm 2, we set $\mu=10^{-4}$, $\lambda=2\mu N$, $\alpha=1.1$, $d=\lceil N/2 \rceil$ and $\epsilon=10^{-3}$. For adaptive selection of $\sigma$, $v$ and $c$ in \eqref{eq:sigmav} and \eqref{eq:sigma}, the parameters are set to $\eta_{\sigma}=0.02, \eta_{v}=\eta_{c}=4, \sigma_{\min}=0.3, v_{\max}=0.2, c_{\min}=0.15$. For rank selection, we set all the elements of the rank to be the same, i.e., $r_1=\ldots =r_N=r$. Then, inspired by \cite{yang2015robust}, the parameter $r$ is determined as $(0.04pI_1I_2)^{1/4}$ and $(0.25pm^2)^{1/4}f^{1/6}$ for global tensor completion and local patch refinement, respectively, where $p$ is the observation rate and $f$ is the number of frames.} For $\ell_{p}$-TTC and $\ell_{p,\epsilon}$-TRC, $p$ is set to $1$. For the other algorithms, the parameters are adjusted so as to achieve the best performance. Further, the parameters are fixed during each simulation. {All algorithms are implemented using MATLAB r2021a on a standard 16-GB memory PC with a 2.6-GHz CPU. The algorithms are run without any acceleration from {(e.g., parallel computation)}.}

\subsection{Color image inpainting}
\label{sc:results:image}
In this section, we verify the robust completion performance on an image inpainting task using the proposed framework, in comparison to other existing tensor ring completion algorithms. Image inpainting takes advantage of the fact that most natural images can be well approximated with their low-rank components, such that filling missing parts of an incomplete image can be regarded as a tensor completion problem.

Test images of size $320\times480\times3$ are selected from the Berkeley Segmentation Dataset \cite{MartinFTM01}. For each image, the pixel value is first normalized to $[0,1]$. Then, $pI_1I_2n$ pixels are selected uniformly at random and set as observed pixels, and the observed pixels are further perturbed with i.i.d. additive noise generated from a given distribution. The image inpainting task is then formulated as a $320\times480\times3$ robust tensor completion problem with {an} observation rate $p$. For TRNN, which favors high-order tensors for better performance \cite{yu2019tensor}, we reshape the tensor to {a} $9$-order tensor of size $4\times4\times4\times5\times4\times4\times5\times6\times3$. For the proposed HQTRC, we reshape the tensor to the same size used for TRNN, while for local patch tensor refinement with HQWTRR, the tensor size is not changed.

\begin{figure}[htbp]
	\centering
	\begin{minipage}[t]{1\linewidth}
		\centering
		\includegraphics[width=1\linewidth]{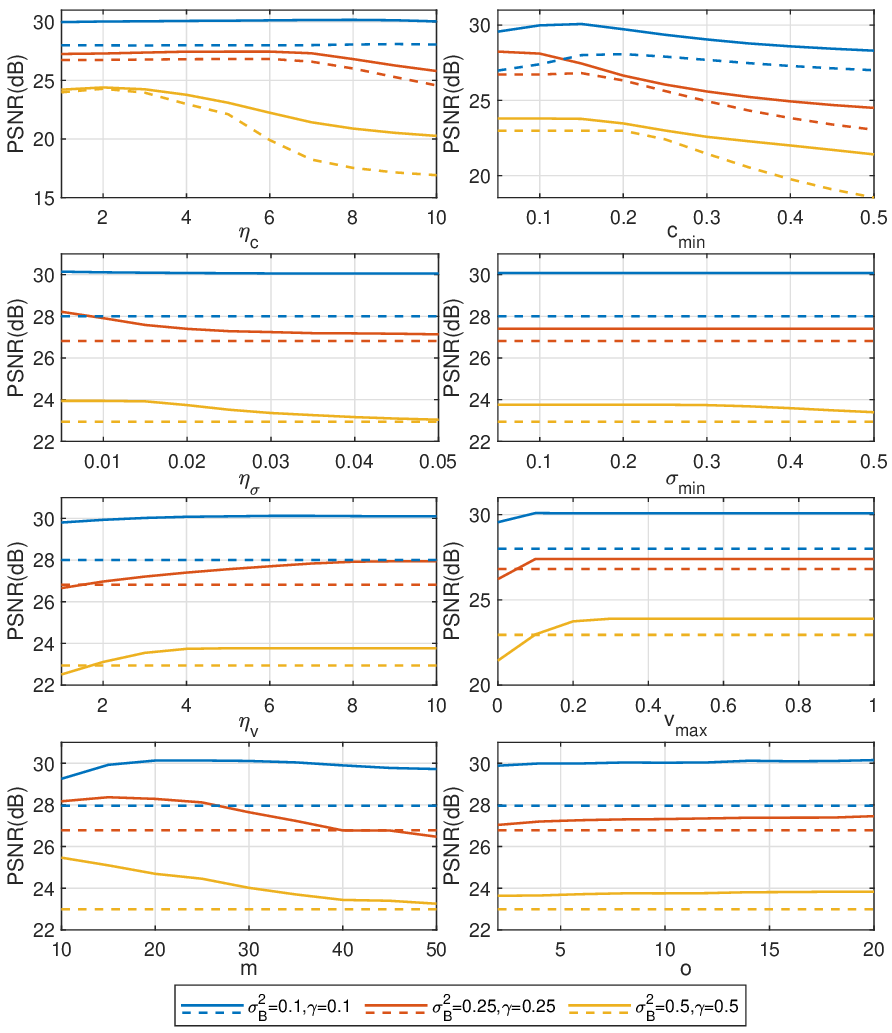}
		\caption{{Curves of average PSNR versus different parameters. Solid lines: C2FRTRC, dotted lines: HQTRC.}}
        \label{fig:ablation_sigma}
	\end{minipage}
	\begin{minipage}[t]{1\linewidth}
		\centering
		\includegraphics[width=1\linewidth]{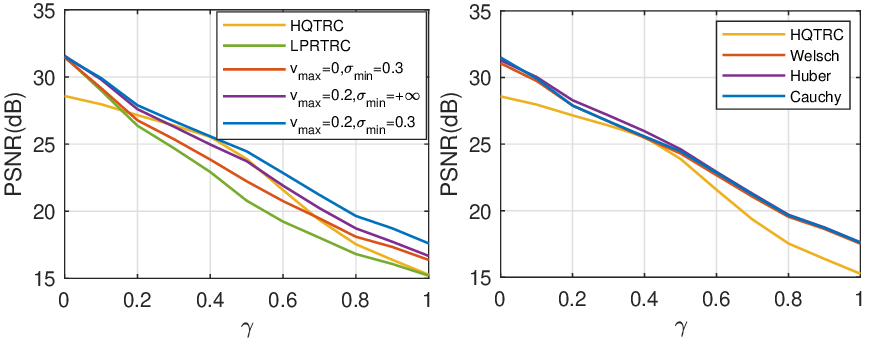}
		\caption{Left: average PSNR versus outlier occurrence probability $\gamma$ under different parameters $v_{\max}$ and $\sigma_{\min}$. Right: average PSNR versus $p$ for different M-estimators.}
		\label{fig:ablation_combine}
	\end{minipage}
\end{figure}

\begin{figure*}[htb]
\centering
\includegraphics[width=0.8\linewidth]{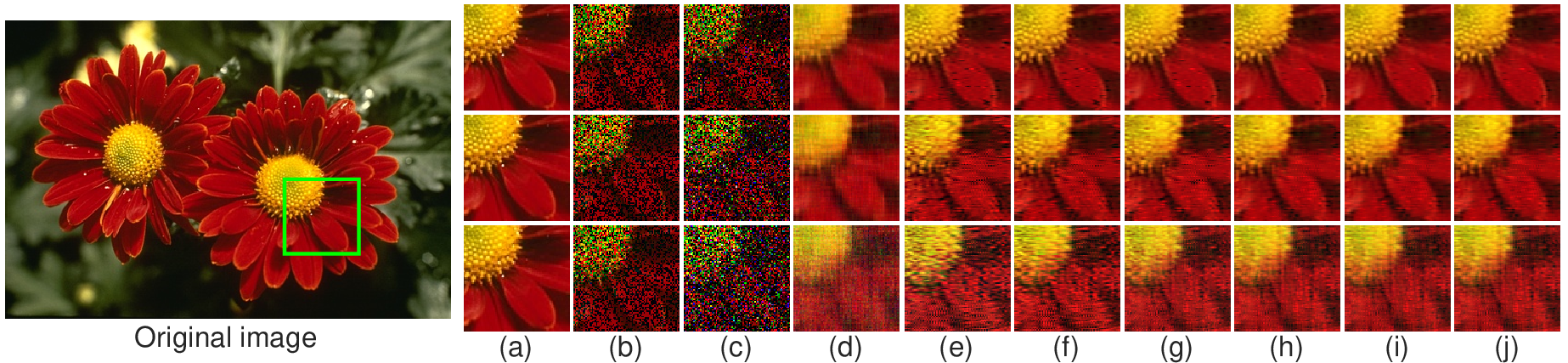}
\caption{Example of the recovered images (partially enlarged) using different parameters. From top to bottom row: $c=0.2,0.5,0.8$. (a) Original partially enlarged image. (b) Noiseless images with missing entries. (c) Noisy image with missing entries (final observed image). (d)-(e) recovered images from HQTRC and LPRTRC. (f)-(g) recovered images from C2FRTRC for different pairs {$v_{\max}$} and {$\sigma_{\min}$}: $\{v_{\max}=0,\sigma_{\min}=0.3\}$ and $\{v_{\max}=0.2,\sigma_{\min}=+\infty\}$. (h)-(j) recovered images from C2FRTRC using different M-estimators: Huber, Welsch and Cauchy.}
\label{fig:ablation_example}
\end{figure*}

\begin{figure*}[htbp]
\centering
\includegraphics[width=1\linewidth]{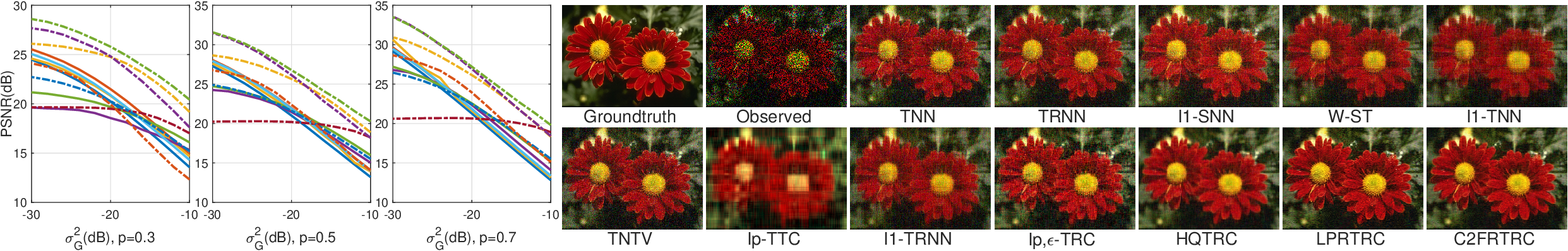}
\includegraphics[width=1\linewidth]{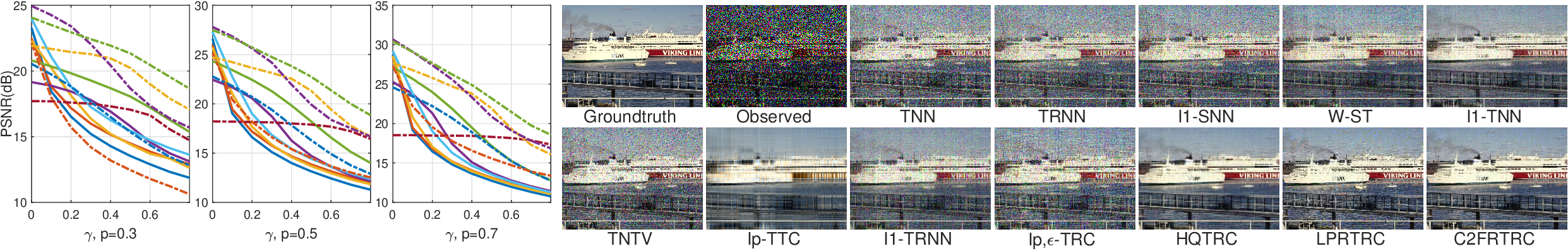}
\includegraphics[width=1\linewidth]{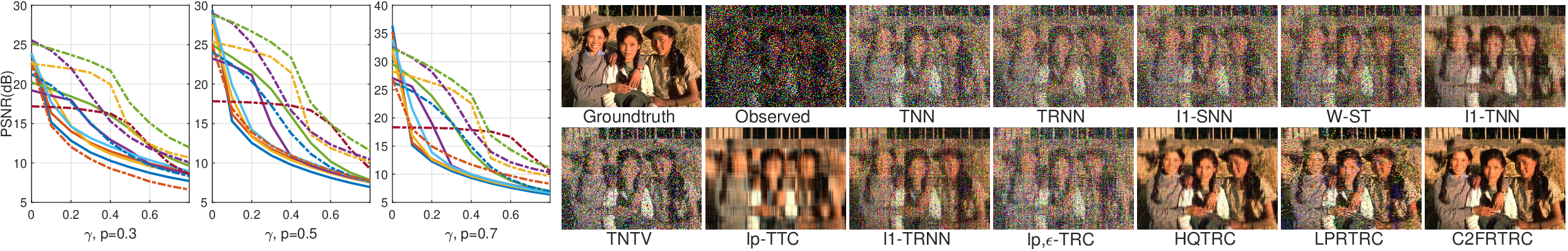}
\includegraphics[width=1\linewidth]{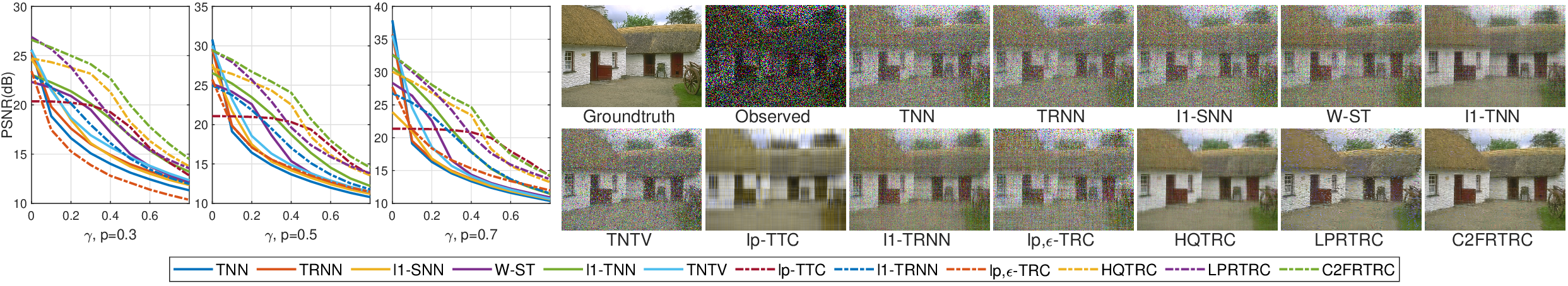}
\caption{{From top to bottom: `flower', `cruise', `girls' and `house'. For each image, (left) Average PSNR versus noise parameter $\gamma/\sigma_G^2$ for different observation rates $p=0.3,0.5,0.7$. (Right) Example of the recovered images ($p=0.5,\gamma=0.4/\sigma_G^2=-20$dB). Best viewed in $\times2$ sized color pdf file.}}
\label{fig:image_example}
\end{figure*}

\subsubsection{Ablation experiment and parameter sensitivity analysis}
We first carry out an ablation experiment and parameter sensitivity analysis on the proposed coarse-to-fine framework and robust tensor ring algorithm. The experiment is carried out on the `flower' image (see Fig. \ref{fig:ablation_example}). The observed pixels are perturbed with additive noise generated from the standard two-component Gaussian mixture model (GMM) with probability density function $(1-\gamma)N(0,{\sigma_A^2})+\gamma N(0,\sigma_B^2)$ . $N(0,{\sigma_A^2})$ represents the general Gaussian noise disturbance with variance ${\sigma_A^2}$, and $N(0,\sigma_B^2)$ with a large variance $\sigma_B^2$ captures the outliers. The variable $\gamma$ controls the occurrence probability of outliers. Unless specified otherwise, the observation rate is set to $p=0.5$, and the GMM noise parameters $\sigma_A^2=0.001, \sigma_B^2=0.25,\gamma=0.5$.

{First, we investigate the parameter sensitivity of the adaptive strategy in \eqref{eq:sigmav} and \eqref{eq:sigma}, along with the patch size $m$ and pixel overlap $o$. Fig. \ref{fig:ablation_sigma} depicts the average PSNR of the recovered image versus $\eta_c$, $\eta_v$, $\eta_{\sigma}$, $v_{\max}$, $\sigma_{\min}$ and $c_{\min}$ under different outlier noise variance $\sigma_B^2$ using HQTRC and C2FRTRC. As shown, C2FRTRC outperforms HQTRC over a wide range of parameters. Specifically, for C2FRTRC, a small value of $\eta_c$, $\eta_{\sigma}$, $c_{\min}$, $\sigma_{\min}$ and $m$ can result in a relatively higher PSNR, while a larger value of $\eta_v$, $v_{\max}$ and $o$ can yield a higher PSNR. One can also observe that when $\eta_c$ and $c_{\min}$ increase to relatively large values, the algorithm cannot properly alleviate the effect of the outliers and the performance degrades.}

{Second, we evaluate the performance of the weighted strategy on local tensor refinement using the global completion result. Apart from LPRTRC which corresponds to the setting $v_{\max}\!=\!0$ and $\sigma_{\min}\!\!=\!\!+\infty$, we also test the results of completion with non-soft weight ($v_{\max}\!=\!0.2$, $\sigma_{\min}\!=\!+\infty$) and without missing entry filling from the global completion result ($v_{\max}\!=\!0$, $\sigma_{\min}\!=\!0.3$). The average PSNR for different outlier occurrence probability $\gamma$ is shown in Fig. \ref{fig:ablation_combine} (left). One can observe that compared with local-only LPRTRC, the completion performance is greatly improved by incorporating the global completion information. The best performance of C2FRTRC verifies that both filling the missing entries with the result of global completion and assigning weights using the soft weighting strategy can improve the completion performance.}

Third, we test performance using different M-estimators. The parameter settings for the Welsch loss function are the same as the Cauchy loss function. For the Huber estimator, the {parameters} $\eta_c$ and $c_{\min}$ are set to $2$ and $0.05$, respectively. The curves of average PSNR with different M-estimators are shown in Fig. \ref{fig:ablation_combine} (right). As shown, the M-estimators yield similar performance, showing the flexibility of the proposed robust method with different selections of M-estimators.

To better illustrate the performance improvement with the proposed framework, we show a visual example of the recovered image using different weight parameters and M-estimators in Fig. \ref{fig:ablation_example}. As can be seen, compared with global tensor completion using HQTRC, the proposed coarse-to-fine framework can improve the texture details, especially in the heavy outlier scenario (2nd and 3rd rows). Further, the global information can also help local patch tensor refinement for accurate estimation of the missing pixels.

\subsubsection{Performance comparison with other algorithms}

In this part, we compare with existing tensor completion algorithms for different noise environments. We use four images (shown in Fig. \ref{fig:image_example}) and add different noise to each image. Specifically, for image `flower', all observed pixels are perturbed with Gaussian noise with zero mean and variance $\sigma_G^2$. For image `cruise', GMM noise with outlier occurrence probability $\gamma$ is added to the observed pixels. For image `girls', $\gamma\times 100\%$ of the observed pixels are perturbed with salt and pepper noise. While for image `house', $\gamma\times 100\%$ of the observed pixels are replaced with random values in the range $[0,1]$.

We investigate the performance on the four images for different settings of the noise parameter $\gamma$ and observation rate $p$. The average PSNR for different algorithms are shown in Fig. \ref{fig:image_example} (left), and an example of the recovered images is shown in Fig. \ref{fig:image_example} (right). It can be observed that the proposed C2FRTRC obtains the overall best performance for different noisy environments. {Specifically, LPRTRC can achieve similar performance to C2FRTRC in Gaussian noise and non-Gaussian noise with small number of outliers ($\gamma\leq 0.2$). While in heavy noise , C2FRTRC is guided by the global prior which can further enhance performance, resulting in better performance than LPRTRC.} One should also notice that when $p=0.7$ and $\gamma=0$ (i.e., noise-free case), TNTV, TRNN and TNN may outperform the proposed method. However, these algorithms suffer from severe performance degradation with a small number of outliers (i.e., $\gamma=0.1$).

\begin{table*}[htbp]
{
\caption{Completion performance comparison for different algorithms on four video sequences with different missing patterns.}
\label{tb:video}
\centering
{
\begin{tabular}{@{}c@{\hspace{3pt}}c@{\hspace{1pt}}c@{\hspace{6pt}}c@{\hspace{9pt}}c@{\hspace{12pt}}c@{\hspace{10pt}}c@{\hspace{10pt}}c@{\hspace{10pt}}c@{\hspace{10pt}}c@{\hspace{10pt}}c@{\hspace{6pt}}c@{\hspace{0pt}}c@{\hspace{4pt}}c@{\hspace{7pt}}c@{\hspace{4pt}}c@{}}
\toprule
Video                    & \multicolumn{2}{l}{\begin{tabular}[c]{@{}c@{}}Noise\\ setting\end{tabular}}                        & Metric  & TNN    & TRNN   & $\ell_1$-SNN & W-ST & $\ell_1$-TNN & TNTV   & $\ell_p$--TTC          & $\ell_1$-TRNN & \multicolumn{1}{c|@{\hspace{4pt}}}{$\ell_{p,\epsilon}$-TRC}       & HQTRC           & LPRTRC          & C2FRTRC         \\ \midrule
\multirow{6}{*}{Tempete} & \multirow{6}{*}{$\gamma$}                                                    & \multirow{2}{*}{0.3} & PSNR   & 11.45  & 11.56  & 11.53  & 12.91  & 22.03  & 17.38  & 17.60           & 18.50   & \multicolumn{1}{c|@{\hspace{4pt}}}{20.57}        & 22.50           & {\ul 25.93}     & \textbf{26.49}  \\
                         &                                                                              &                      & SSIM   & 0.3122 & 0.3177 & 0.3148 & 0.3371 & 0.7218 & 0.5338 & 0.5193          & 0.5839  & \multicolumn{1}{c|@{\hspace{4pt}}}{0.6417}       & 0.8088          & {\ul 0.9115}    & \textbf{0.9194} \\
                         &                                                                              & \multirow{2}{*}{0.5} & PSNR   & 9.29   & 9.35   & 9.44   & 10.60  & 14.60  & 12.26  & 17.64           & 12.42   & \multicolumn{1}{c|@{\hspace{4pt}}}{20.41}        & 22.59           & {\ul 24.62}     & \textbf{25.33}  \\
                         &                                                                              &                      & SSIM   & 0.2215 & 0.2265 & 0.2252 & 0.2403 & 0.4050 & 0.3193 & 0.5179          & 0.3227  & \multicolumn{1}{c|@{\hspace{4pt}}}{0.6267}       & 0.7797          & {\ul 0.8559}    & \textbf{0.8818} \\
                         &                                                                              & \multirow{2}{*}{0.7} & PSNR   & 7.78   & 7.96   & 7.93   & 8.98   & 10.94  & 9.66   & 17.76           & 9.55    & \multicolumn{1}{c|@{\hspace{4pt}}}{20.13}        & {\ul 20.88}     & 19.02           & \textbf{22.19}  \\
                         &                                                                              &                      & SSIM   & 0.1632 & 0.1661 & 0.1634 & 0.1627 & 0.2450 & 0.2030 & 0.5144          & 0.2013  & \multicolumn{1}{c|@{\hspace{4pt}}}{0.6163}       & {\ul 0.6440}    & 0.5636          & \textbf{0.7082} \\ \midrule
\multirow{6}{*}{Stefan}  & \multirow{6}{*}{$\gamma$}                                                    & \multirow{2}{*}{0.3} & PSNR   & 11.63  & 12.14  & 11.71  & 8.95   & 18.87  & 13.51  & 17.18           & 12.72   & \multicolumn{1}{c|@{\hspace{4pt}}}{17.89}        & 19.82           & {\ul 22.32}     & \textbf{22.48}  \\
                         &                                                                              &                      & SSIM   & 0.2079 & 0.2245 & 0.2223 & 0.1449 & 0.6037 & 0.2949 & 0.5378          & 0.2472  & \multicolumn{1}{c|@{\hspace{4pt}}}{0.5061}       & 0.6819          & \textbf{0.8117} & {\ul 0.8025}    \\
                         &                                                                              & \multirow{2}{*}{0.5} & PSNR   & 9.51   & 10.11  & 9.55   & 7.59   & 11.50  & 10.28  & {\ul 17.13}     & 10.03   & \multicolumn{1}{c|@{\hspace{4pt}}}{16.73}        & 16.11           & 15.55           & \textbf{19.95}  \\
                         &                                                                              &                      & SSIM   & 0.1112 & 0.1254 & 0.1091 & 0.0630 & 0.1818 & 0.1391 & {\ul 0.5375}    & 0.1224  & \multicolumn{1}{c|@{\hspace{4pt}}}{0.4090}       & 0.3309          & 0.3273          & \textbf{0.6672} \\
                         &                                                                              & \multirow{2}{*}{0.7} & PSNR   & 8.01   & 8.72   & 8.24   & 6.71   & 8.94   & 8.65   & {\ul 16.89}     & 8.56    & \multicolumn{1}{c|@{\hspace{4pt}}}{15.10}        & 14.57           & 13.70           & \textbf{17.34}  \\
                         &                                                                              &                      & SSIM   & 0.0570 & 0.0615 & 0.0608 & 0.0509 & 0.0718 & 0.0667 & \textbf{0.5251} & 0.0610  & \multicolumn{1}{c|@{\hspace{4pt}}}{0.1917}       & 0.1543          & 0.1620          & {\ul 0.4598}    \\ \midrule
\multirow{6}{*}{Foreman} & \multirow{6}{*}{$\gamma$}                                                    & \multirow{2}{*}{0.3} & PSNR   & 11.40  & 11.55  & 11.51  & 11.11  & 21.96  & 18.79  & 18.21           & 19.13   & \multicolumn{1}{c|@{\hspace{4pt}}}{22.45}        & 25.58           & {\ul 29.52}     & \textbf{30.71}  \\
                         &                                                                              &                      & SSIM   & 0.1955 & 0.2137 & 0.2128 & 0.1985 & 0.5887 & 0.4758 & 0.6704          & 0.5022  & \multicolumn{1}{c|@{\hspace{4pt}}}{0.7208}       & 0.8781          & {\ul 0.9270}    & \textbf{0.9275} \\
                         &                                                                              & \multirow{2}{*}{0.5} & PSNR   & 9.24   & 9.39   & 8.94   & 9.35   & 14.61  & 12.54  & 18.26           & 12.46   & \multicolumn{1}{c|@{\hspace{4pt}}}{22.08}        & 25.37           & {\ul 27.67}     & \textbf{29.47}  \\
                         &                                                                              &                      & SSIM   & 0.1354 & 0.1495 & 0.1457 & 0.1378 & 0.2815 & 0.2267 & 0.6674          & 0.2274  & \multicolumn{1}{c|@{\hspace{4pt}}}{0.7021}       & 0.8433          & {\ul 0.8750}    & \textbf{0.9117} \\
                         &                                                                              & \multirow{2}{*}{0.7} & PSNR   & 7.85   & 7.93   & 7.87   & 7.95   & 10.99  & 9.69   & 18.37           & 9.50    & \multicolumn{1}{c|@{\hspace{4pt}}}{21.85}        & {\ul 22.06}     & 19.51           & \textbf{23.99}  \\
                         &                                                                              &                      & SSIM   & 0.1007 & 0.1136 & 0.1138 & 0.1023 & 0.1597 & 0.1432 & 0.6593          & 0.1397  & \multicolumn{1}{c|@{\hspace{4pt}}}{{\ul 0.6927}} & 0.6496          & 0.5103          & \textbf{0.7356} \\ \midrule
\multirow{6}{*}{Bus}     & \multirow{6}{*}{\begin{tabular}[c]{@{}c@{}}$\sigma_G^2$\\ (dB)\end{tabular}} & \multirow{2}{*}{-20} & PSNR   & 20.09  & 20.17  & 19.95  & 19.53  & 21.40  & 21.07  & 16.54           & 20.51   & \multicolumn{1}{c|@{\hspace{4pt}}}{19.34}        & 21.66           & \textbf{25.20}  & {\ul 24.06}     \\
                         &                                                                              &                      & SSIM   & 0.5108 & 0.5333 & 0.5253 & 0.5049 & 0.5446 & 0.5597 & 0.3154          & 0.5400  & \multicolumn{1}{c|@{\hspace{4pt}}}{0.4045}       & 0.6226          & \textbf{0.7411} & {\ul 0.7221}    \\
                         &                                                                              & \multirow{2}{*}{-15} & PSNR   & 15.42  & 15.89  & 15.66  & 13.78  & 17.36  & 17.05  & 16.63           & 16.34   & \multicolumn{1}{c|@{\hspace{4pt}}}{19.22}        & 21.09           & \textbf{22.59}  & {\ul 22.03}     \\
                         &                                                                              &                      & SSIM   & 0.3109 & 0.3343 & 0.3265 & 0.2693 & 0.3514 & 0.3674 & 0.2928          & 0.3425  & \multicolumn{1}{c|@{\hspace{4pt}}}{0.3862}       & 0.5511          & \textbf{0.5884} & {\ul 0.5878}    \\
                         &                                                                              & \multirow{2}{*}{-10} & PSNR   & 10.46  & 10.70  & 10.67  & 10.65  & 12.29  & 11.67  & 16.61           & 11.28   & \multicolumn{1}{c|@{\hspace{4pt}}}{19.05}        & {\ul 20.01}     & 18.74           & \textbf{20.08}  \\
                         &                                                                              &                      & SSIM   & 0.1662 & 0.1734 & 0.1710 & 0.1560 & 0.1970 & 0.1938 & 0.2720          & 0.1838  & \multicolumn{1}{c|@{\hspace{4pt}}}{0.3649}                            & \textbf{0.4384} & 0.3919          & {\ul 0.4350}    \\ \bottomrule
\end{tabular}%
}
}
\end{table*}

\subsection{Video completion}

\begin{figure*}[htbp]
\centering
\includegraphics[width=1\linewidth]{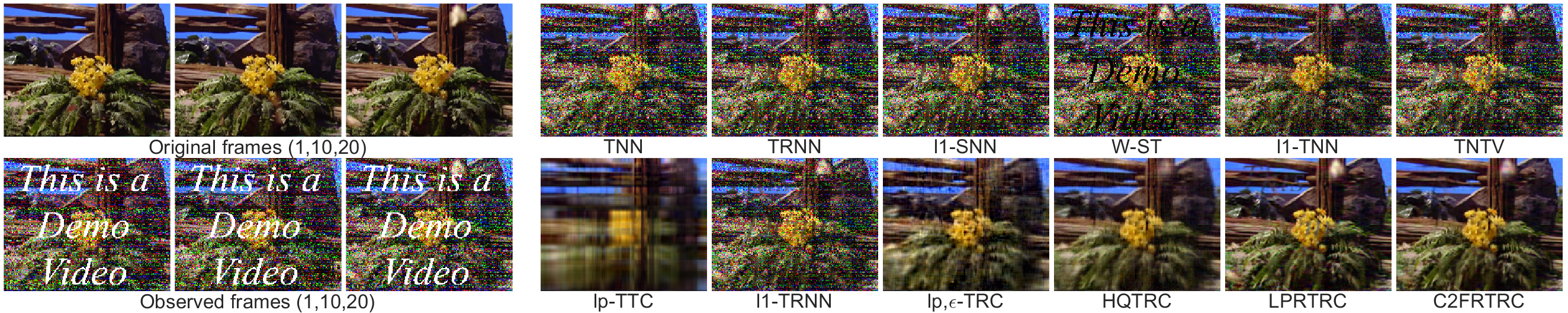}
\includegraphics[width=1\linewidth]{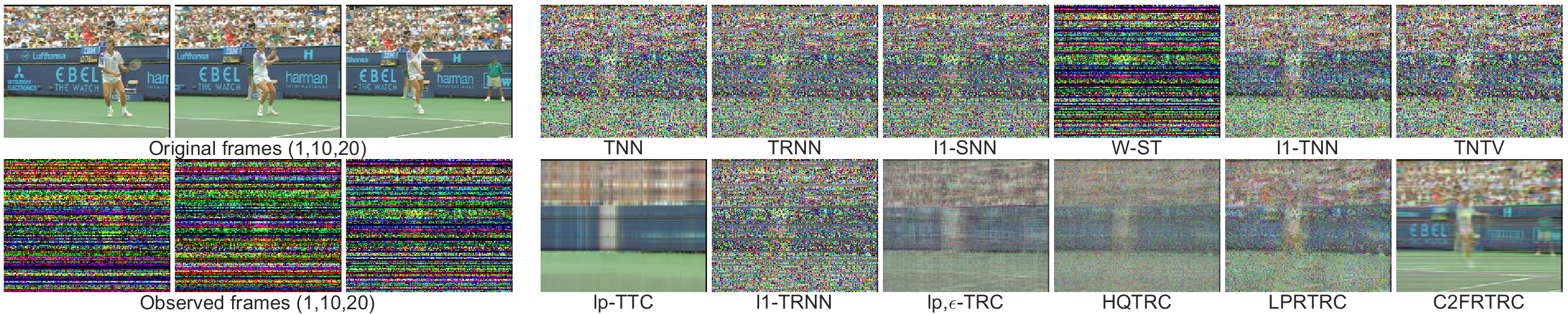}
\includegraphics[width=1\linewidth]{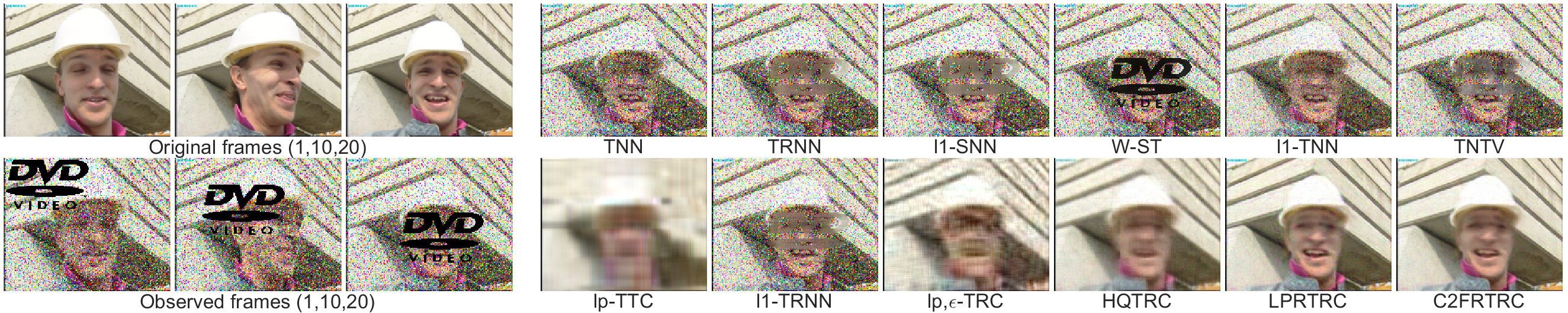}
\includegraphics[width=1\linewidth]{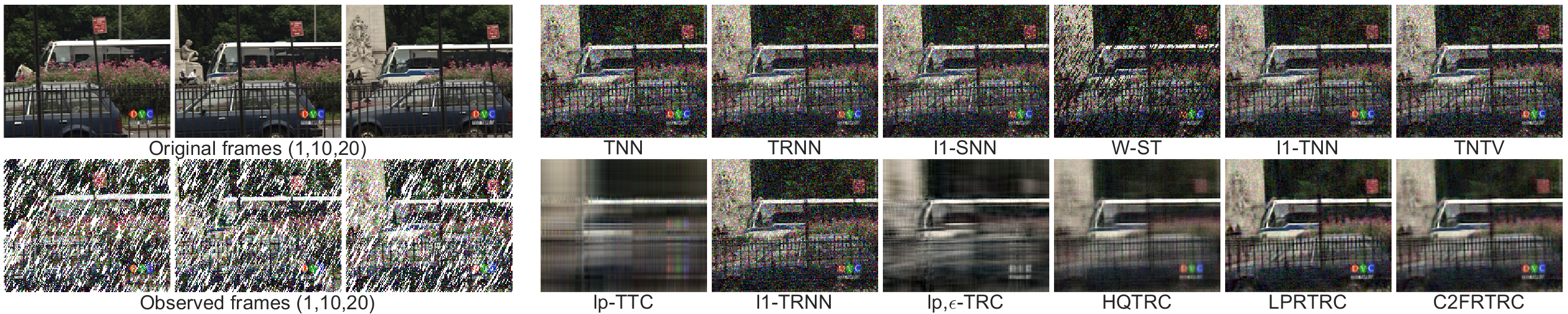}
\caption{From top to bottom: video `tempete', `stefan', `foreman' and `bus'. For each video: (left) The original frames (frame number: $1,10,20$) and corresponding observed noisy frames with missing pixels ($\gamma=0.5/\sigma_G^2=-15$dB); (Right) Recovered $20$-th frame using different algorithms. Best viewed in $\times2$ sized color pdf file.}
\label{fig:video_example}
\end{figure*}

In this part, we compare the performance of the proposed method with existing robust tensor completion algorithms in a video completion task. The completion performance is evaluated using four color video fragments from the YUV dataset\footnote{http://trace.eas.asu.edu/yuv/}. Some frames of the original videos are shown in Fig. \ref{fig:video_example}. For each video, a sequence of 30 frames is selected, and each frame is resized to $144\times180$ to obtain a tensor of size $144\times180\times3\times30$. Similar to the previous section, a tensor with noisy and missing (partially observed) entries is generated by selecting a fraction of pixels as observed pixels and then adding i.i.d. noise from a given distribution to the observed pixels. For TRNN and HQTRC, the observed tensor is reshaped to {an} $11$-order tensor of size $3\times3\times4\times4\times3\times3\times4\times5\times3\times5\times6$. For $\ell_1$-SNN, W-ST $\ell_1$-TNN and TNTV, we reshape the tensor to a $3$-order tensor of size $144\times180\times90$.

We apply different types of missing patterns and noise distributions to each video fragment. In particular, for video `tempete', a fixed sentence is masked on all frames so that the video contains a `watermark', and $\gamma\times 100\%$ of the rows in each frame are perturbed with outliers generated from a Gaussian distribution with zero mean and variance $0.25$. For video 'stefan', $60\%$ of the rows are randomly and uniformly selected as the observed rows, and the observed data is perturbed with salt and pepper noise with probability $\gamma\times100\%$. For video `foreman', a watermark moving from the top-left to the bottom right of the video is added as the missing pattern, then GMM noise with $\sigma_A^2=0.001,\sigma_B^2=0.25$ and outlier occurrence probability $\gamma$ is added to the observed pixels. Finally, for video `bus', we use a time-variant missing pattern to simulate the effect of the raindrop, and a Gaussian noise with zero mean and variance $\sigma_G^2$ is added to the observed data. Representatives of the observed noisy frames are shown in Fig. \ref{fig:video_example}.

Table \ref{tb:video} shows the average PSNR {and SSIM} for different algorithms on four video fragments in different noise environments. {The SSIM of a video is computed as the average SSIM across frames. As shown, C2FRTRC achieves the overall best performance, and LPRTRC yields the second-best performance. Specifically, C2FRTRC achieves the highest PSNR/SSIM in most cases. In Gaussian noise environments with $\sigma_G^2\leq-15$dB, LPRTRC outperforms C2FRTRC.} Fig. \ref{fig:video_example} shows an example of the recovered frames from the four fragments in heavy noise environments. As can be seen, only the proposed C2FRTRC successfully recovers the frames of all videos. Further, similar to image completion, C2FRTRC yields the best visual results having the most clean and detailed texture.

\begin{figure}[htbp]
\centering
\includegraphics[width=1\linewidth]{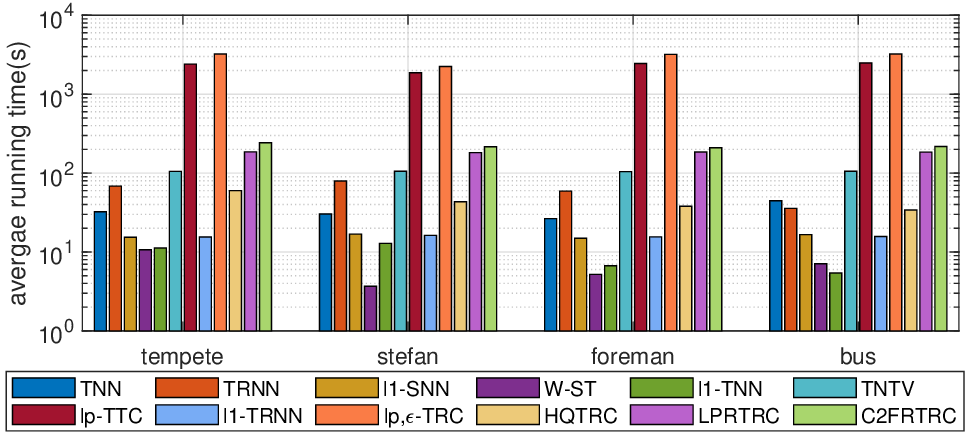}
\caption{{Average running times of four videos corresponding to Table \ref{tb:video}.}}
\label{fig:video_time}
\end{figure}

{Fig. \ref{fig:video_time} reports the average running time of all algorithms on the four videos in Table \ref{tb:video}. It can be seen that the coarse completion HQTRC incurs a time cost  similar to $\ell_1$-based algorithms $\ell_1$-SNN, $\ell_1$-TNN and $\ell$-TRNN. For C2FRTRC, the time cost is higher due to the extra refinement step for local patch tensors. As refinement for each patch tensor is independent, the time cost of C2FRTRC could be reduced using parallel computation. Although the local fine stage incurs the additional time cost, the performance improvement is significant.}

\section{Conclusion}
\label{sec:conc}
We proposed a novel two-stage coarse-to-fine tensor completion framework for robust visual data completion. A global coarse completion stage is first performed, whereby most of the outliers are identified. Then, guided by the result of global completion, a local patch refinement process is applied by performing robust tensor recovery incorporating both local and global information. {Further, a new M-estimator-based tensor ring recovery algorithm using HQ approach is proposed, which can accurately complete and recover the tensor in the presence of a large number of outliers. Numerical experiments on image and video completion in various noise environments demonstrate the advantage of incorporating global coarse completion with local patch refinement. The results also demonstrate that the proposed method can outperform existing state-of-the-art robust tensor completion algorithms, especially in heavy noise settings. }  

\appendices

\section{Proof of Theorem \ref{thm:HQTRC}}
{
From \eqref{eq:trnn_x} and \eqref{eq:trnn_g}, we have
\begin{equation}
\begin{aligned}
\mathcal{L}^{t+1}&\!=\!\frac{1}{N}\sum_{k=1}^{N}\left(\mathcal{Z}^{(k),t+1}+\frac{1}{\mu}\mathcal{G}^{(k),t}\right)\\
&\!=\!\frac{1}{N}\sum_{k=1}^{N}\!\left(\mathcal{Z}^{(k),t+1}\!+\!\frac{1}{\mu}(\mathcal{G}^{(k),t+1}\!-\!\mu(\mathcal{Z}^{(k),t+1}\!-\!\mathcal{X}^{t+1}))\!\right)\\
&\!=\!\mathcal{X}^{t+1}+\frac{1}{\mu N}\sum_{k=1}^{N}\mathcal{G}^{(k),t+1}\:.
\end{aligned}
\tag{S.1}
\end{equation}
Substituting (S.1) into \eqref{eq:trnn_x} we get
\begin{equation}
\begin{aligned}
\mathcal{G}_s^{t+1}={\lambda\mathcal{V}^{t+1}}\circ(\mathcal{X}^{t+1}-\mathcal{M})
\end{aligned}
\tag{S.2}
\end{equation}
where $\mathcal{G}_s^{t+1}=\sum_{k=1}^{N}\mathcal{G}^{(k),t+1}$ and $\mathcal{V}^{t+1}=\mathcal{W}\circ\mathcal{Q}^{t+1}$. 
Since $\mathcal{W}_{i_1\ldots i_N}=0$ for $(i_1,\ldots,i_N) \notin \Omega$, we get that $(\mathcal{G}_s^{t+1})_{i_1\ldots i_N}=0$ for $(i_1,\ldots,i_N) \notin \Omega$. Next, we will show that $\lim_{c\rightarrow0}(\mathcal{G}_s^{t+1})_{i_1,\ldots,i_N}=0$ for $(i_1,\ldots,i_N)\in\Omega$.

Here, we use the Welsch function as an example. By applying the Welsch function to the second term on the right hand side (RHS) of \eqref{eq:trnn_x}, we have that for all $(i_1\ldots i_N)\in\Omega$
\begin{equation}
\begin{aligned}
&\mathcal{Q}^{t+1}_{i_1\ldots i_N}(\mathcal{M}_{i_1\ldots i_N}-\mathcal{L}^{t+1}_{i_1\ldots i_N})\\
&=\exp\left({-\frac{(\mathcal{M}_{i_1\ldots i_N}-\mathcal{X}^t_{i_1\ldots i_N})^2}{2c^2}}\right)(\mathcal{M}_{i_1\ldots i_N}-\mathcal{X}^{t+1}_{i_1\ldots i_N}\\
&~~~~-\frac{1}{\mu N}(\mathcal{G}_s^{t+1})_{i_1\ldots i_N})\:.
\end{aligned}
\tag{S.3}
\end{equation}
Based on the assumption that $\|\mathcal{X}^{t+1}-\mathcal{X}^{t}\|_F^2<\infty$, we get that $|\mathcal{X}^{t+1}_{i_1\ldots i_N}-\mathcal{X}^{t}_{i_1\ldots i_N}|\leq P$, where $P$ is some finite value. Thus, we have
\begin{equation}
\begin{aligned}
a\leq\mathcal{Q}^{t+1}_{i_1\ldots i_N}(\mathcal{M}_{i_1\ldots i_N}-\mathcal{L}^{t+1}_{i_1\ldots i_N})\leq b
\end{aligned}
\tag{S.4}
\end{equation}
where
$$
a\!=\exp\left({-\frac{(\mathcal{E}^{t}_{i_1\ldots i_N})^2}{2c^2}}\right)(\mathcal{E}^{t}_{i_1\ldots i_N}-\frac{1}{\mu N}(\mathcal{G}_s^{t+1})_{i_1\ldots i_N}-P)
$$
$$
b\!=\exp\left({-\frac{(\mathcal{E}^{t}_{i_1\ldots i_N})^2}{2c^2}}\right)(\mathcal{E}^{t}_{i_1\ldots i_N}-\frac{1}{\mu N}(\mathcal{G}_s^{t+1})_{i_1\ldots i_N}+P)
$$
where $\mathcal{E}^{t}_{i_1\ldots i_N}=\mathcal{M}_{i_1\ldots i_N}-\mathcal{X}^{t}_{i_1\ldots i_N}$. For the Welsch function $f(x)=c^2(1-\exp(-x^2/(2c^2)))$, $x\in\mathbb{R}$, $f'(x)=x\exp(-\frac{x^2}{2c^2})\in [-ce^{-0.5},ce^{-0.5}]$, hence $a$ and $b$ are bounded for any values of $\mathcal{X}^{t}_{i_1\ldots i_N}$. It can be also observed that both $a$ and $b$ are $0$ when $c\rightarrow0$ and $\mathcal{E}^{t}_{i_1\ldots i_N}\neq0$. Therefore, for $\mathcal{E}^{t}_{i_1\ldots i_N}\neq0$, from (S.4) we have that $\lim_{c\rightarrow0}\mathcal{Q}^{t+1}_{i_1\ldots i_N}(\mathcal{M}_{i_1\ldots i_N}-\mathcal{L}^{t+1}_{i_1\ldots i_N})=0$. Then, using (S.2) one can further obtain that $\lim_{c\rightarrow0}(\mathcal{G}_s^{t+1})_{i_1\ldots i_N}=0$. 

The key point of the above analysis is the boundedness of $f'(x)$. Since $f'(x)$ is also bounded for the Cauchy and Huber functions, a similar result can be derived.

Combining the results above, we conclude that $\lim_{c\rightarrow0}(\mathcal{G}_s^{t+1})_{i_1\ldots i_N}=0$ for $\{({i_1,\ldots, i_N}):\mathcal{E}^{t}_{i_1\ldots i_N}\neq0\}$. Further, since $\{\mathcal{G}^{(k),t}\}$ converges to $\mathcal{C}$, we get that $\lim_{c\rightarrow0}\mathcal{C}_{i_1\ldots i_N}=0$ for $\{({i_1,\ldots,i_N}): \lim_{t\rightarrow\infty}\mathcal{E}^{t}_{i_1\ldots i_N}\neq0\}$. Moreover, for the indices $({i_1,\ldots,i_N})$ that $\lim_{t\rightarrow\infty}\mathcal{E}^{t}_{i_1\ldots i_N}=0$, we have $\lim_{t\rightarrow\infty}\mathcal{X}^{t}_{i_1\ldots i_N}=\mathcal{M}_{i_1\ldots i_N}$, and from (S.2) one can obtain $\mathcal{C}_{i_1\ldots i_N}=0$. Therefore, we get that $\lim_{c\rightarrow0}\mathcal{C}=0$.

We also conclude from \eqref{eq:trnn_g} that $\left\{\mathcal{Z}^{(k),t}-\mathcal{X}^t\right\}$ converges to $0$ for all $k=1,\ldots,N$. Therefore, from \eqref{eq:trnn_z_cf} we have 
\begin{equation}
\begin{aligned}
\lim_{c\rightarrow0}\lim_{t\rightarrow\infty}\mathbf{X}^{t+1}_{\langle k,d\rangle}&\!=\!\Pi_{r_{kd}}(\lim_{c\rightarrow0}\lim_{t\rightarrow\infty}\mathbf{X}^t_{\langle k,d\rangle}\!-\!\frac{1}{\mu}\lim_{c\rightarrow0}\lim_{t\rightarrow\infty}\mathbf{G}_{\langle k,d\rangle}^{(k),t})\\
&=\Pi_{r_{kd}}(\lim_{c\rightarrow0}\lim_{t\rightarrow\infty}\mathbf{X}^t_{\langle k,d\rangle})\:.
\end{aligned}
\tag{S.5}
\end{equation}
From the property of the truncated SVD, we have that, $\lim_{c\rightarrow0}\lim_{t\rightarrow\infty}\mathcal{X}^{t+1}\!=\!\lim_{c\rightarrow0}\lim_{t\rightarrow\infty}\mathcal{X}^{t}$. Therefore, we conclude that as $c$ decreases to $0$, $\{\mathcal{X}^t\}$ converges.}

\bibliographystyle{IEEEtran}
\bibliography{main}
\end{document}